\documentclass[11pt]{article}

\usepackage[final]{acl}

\usepackage{times}
\usepackage{latexsym}

\usepackage[T1]{fontenc}
\usepackage[utf8]{inputenc}

\usepackage{microtype}

\usepackage{inconsolata}

\usepackage{graphicx}
\usepackage{booktabs}
\usepackage{pifont}
\definecolor{darkgreen}{rgb}{0.0, 0.5, 0.0} 
\definecolor{verylightgray}{rgb}{0.97, 0.97, 0.97}
\definecolor{darkorange}{rgb}{0.8, 0.4, 0.0}

\newcommand{\cmark}{\textcolor{darkgreen}{\scalebox{1}[1.0]{\ding{51}}}}
\newcommand{\xmark}{\textcolor{red}{\ding{55}}}
\usepackage[table]{xcolor}
\usepackage{multirow}
\usepackage{amsmath}
\usepackage{algorithm}
\usepackage{algpseudocode}
\usepackage{mdframed}
\usepackage{graphicx}
\usepackage{cuted}
\usepackage{caption} 
\usepackage[most]{tcolorbox}
\usepackage{booktabs}
\usepackage{tabularx}
\usepackage{enumitem} 
\usepackage{listings}
\usepackage{subcaption}
\usepackage{dblfloatfix}
\usepackage{amsfonts}

\tcbuselibrary{listings,breakable}

\newtcolorbox{promptbox}{
  colback=gray!5,
  colframe=black!70,
  boxrule=0.5pt,
  arc=4pt,
  left=6pt,
  right=6pt,
  top=6pt,
  bottom=6pt
}

\tcbuselibrary{breakable}

\title{MedRoundsQA: A Persona and Difficulty Aware Evaluation for \\Multi-Turn Medical Consultations}

\author{
 \textbf{Youssef Mohamed}\textsuperscript{1}, \textbf{Ahmed Heakl}\textsuperscript{1}, \textbf{Qinrong Cui}\textsuperscript{1}, \textbf{Junhong Liang}\textsuperscript{1},\\
 \textbf{Rafiq Ali}\textsuperscript{1}, \textbf{Bdour Babillie}\textsuperscript{1}, \textbf{Nazira Dunbayeva}\textsuperscript{1}, \textbf{Lang Gao}\textsuperscript{1}, \textbf{Omar Hussein}\textsuperscript{2},\\
 \textbf{Ahmed Nada}\textsuperscript{2}, \textbf{Ahmed Mohamed Magdy Mohamed}\textsuperscript{3}, \textbf{Jinghui Liu}\textsuperscript{4},\\
 \textbf{Salman Khan}\textsuperscript{1}, \textbf{Imran Razzak}\textsuperscript{1}, \textbf{Yuxia Wang}\textsuperscript{5}, \textbf{Xiuying Chen}\textsuperscript{1,\dag}\\
 \textsuperscript{1}MBZUAI, \textsuperscript{2}Cairo University, \textsuperscript{3}Ain Shams University, \textsuperscript{4}CSIRO, \textsuperscript{5}INSAIT, Sofia University “St. Kliment Ohridski”\\
 \texttt{\{youssef.mohamed,xiuying.chen\}@mbzuai.ac.ae}
}

\begin{document}
\maketitle

\begingroup
\renewcommand\thefootnote{}
\footnotetext{
\textsuperscript{\dag} Corresponding author.
\quad
Code and data: \url{https://github.com/youssefkhalil320/MedRoundsQA}
}
\addtocounter{footnote}{-1}
\endgroup

\begin{abstract}
Medical benchmarks are dominated by single-turn, multiple-choice
clinical cases that poorly reflect real consultations. Practically,
clinicians elicit evidence interactively and patient communication
varies widely. We introduce \textsc{MedRoundsQA}, a multi-turn
diagnostic benchmark derived from 1{,}387 board-exam cases across
17 specialties. Each case is converted into a structured 24-slot
clinical record, and then instantiated as controlled doctor-patient
dual-agent dialogues under varying patient personas, with the
underlying clinical content held fixed.
We further classify cases by difficulty using model-based uncertainty
to enable easy-to-hard analysis. Evaluations of fifteen LLM doctor
agents show that \emph{(i)} moving from a single-turn diagnosis on the
standardized records to multi-turn consultations causes large
degradations of roughly 13--39 points; \emph{(ii)} more turns reliably
improves question relevance, but diagnostic accuracy exhibits
diminishing returns and typically plateaus after 6--12 turns; and
\emph{(iii)} patient persona differences can shift diagnosis accuracy
by about 7--8 points (lowest to highest education), highlighting
equity risks that single-turn benchmarks miss.
\end{abstract}

\section{Introduction}

Medical diagnosis is inherently interactive: clinicians gather information incrementally through targeted questioning, and the quality of that information depends on how patients communicate. Yet current benchmarks ignore both properties, presenting models with complete clinical vignettes for single-turn diagnosis — bypassing the sequential reasoning and patient-facing communication that define real consultations.

This disconnect matters. LLMs are already deployed in clinical workflows: 20\% of UK GPs use tools like ChatGPT in practice~\citep{blease2024genai}, and one in six US adults consult AI chatbots for health information monthly~\citep{healthline2025chatgpt}. Failures when information arrives gradually or patients communicate imprecisely carry consequences beyond benchmark scores.

\begin{figure*}[t!]
  \centering
  \includegraphics[width=0.92\textwidth, trim={15 10 15 10}, clip]{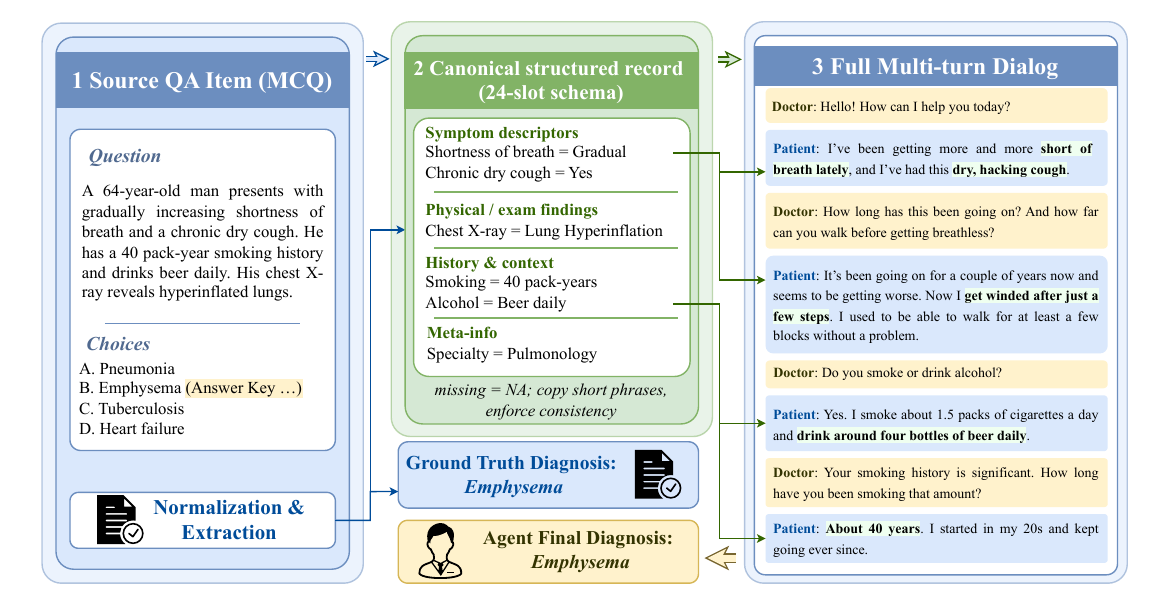}
  \caption{\textbf{From static vignettes to interactive consultations.}
  A board-exam MCQ (left) is normalized into a 24-slot clinical schema
  (center) and used to generate grounded multi-turn dialogues (right).
  This design decouples clinical content from surface realization,
  enabling controlled evaluation of how patient communication style
  affects diagnostic accuracy on identical underlying cases.}
  \label{fig:dialoug}
\end{figure*}

Existing resources fall into two categories, neither sufficient. Single-turn benchmarks (MedQA, MedMCQA, MultiMedQA) provide exam-quality ground truth but treat each case as a static input~\citep{jin2020medqa,pal2022medmcqa,singhal2023largelanguage}. Dialogue corpora (MedDialog, MedDG) capture multi-turn structure but lack controlled diagnostic ground truth~\citep{he2020meddialog,liu2022meddg}. Agent-based benchmarks (AgentClinic, MediQ, AI Hospital) introduce interactive evaluation but vary patients across different cases, making persona effects inseparable from case difficulty~\citep{schmidgall2024agentclinic,li2024mediq,fan2024aihospital}.

\textbf{Key missing capability: counterfactual control.} Attributing performance differences to communication requires instantiating the same clinical case under multiple patient personas with clinical truth held fixed, enabling causal measurement of robustness and equity gaps. Table~\ref{tab:comparison} summarizes how existing resources fall short.

\begin{table*}[t]
\centering
\small
\setlength{\tabcolsep}{3.5pt}
\rowcolors{2}{white}{gray!10}
\resizebox{\linewidth}{!}{
\begin{tabular}{lllcccccccccl}
\toprule
\textbf{Benchmark} & \textbf{Type} & \textbf{Source} & \textbf{Size} &
\textbf{MT} & \textbf{MA} & \textbf{Persona} &
\textbf{DF} & \textbf{Schema} &
\textbf{CF} & \textbf{ST$\leftrightarrow$MT} & \textbf{FC} & \textbf{Lang.} \\
\midrule
\multicolumn{13}{l}{\textit{Single-turn QA benchmarks}} \\
MedQA~\citep{jin2020medqa}                 & QA        & Board exams     & 12.7k & \xmark & \xmark & \xmark & \xmark & \xmark & \xmark & \xmark & \xmark & EN/ZH \\
MedMCQA~\citep{pal2022medmcqa}             & QA        & Entrance exams  & 194k  & \xmark & \xmark & \xmark & \xmark & \xmark & \xmark & \xmark & \xmark & EN \\
MultiMedQA~\citep{singhal2023largelanguage}& QA        & Mixed           & 7 sets& \xmark & \xmark & \xmark & \xmark & \xmark & \xmark & \xmark & \xmark & EN \\
\midrule
\multicolumn{13}{l}{\textit{Dialogue corpora (training-focused)}} \\
MedDialog-EN~\citep{he2020meddialog}       & Corpus    & Telemedicine    & 300k  & \cmark & \xmark & \xmark & \xmark & \xmark & \xmark & \xmark & \xmark & EN \\
MedDG~\citep{liu2022meddg}                 & Corpus    & Telemedicine    & 17k   & \cmark & \xmark & \xmark & \xmark & \textit{Partial} & \xmark & \xmark & \xmark & ZH \\
\midrule
\multicolumn{13}{l}{\textit{Multi-agent dialogue benchmarks}} \\
MediQ~\citep{li2024mediq}                  & Benchmark & Curated         & 1.2k  & \cmark & \cmark & \xmark           & \xmark & \xmark & \xmark & \xmark & \xmark & EN \\
AgentClinic~\citep{schmidgall2024agentclinic} & Benchmark & Mixed       & 457   & \cmark & \cmark & \xmark           & \xmark & \xmark & \xmark & \xmark & \xmark & EN \\
AI Hospital~\citep{fan2024aihospital}      & Benchmark & Curated         & 506   & \cmark & \cmark & \textit{Temper.} & \xmark & \xmark & \xmark & \xmark & \xmark & ZH \\
Dr.APP~\citep{zhu2025drapp}                & Benchmark & Curated         & 1.5k  & \cmark & \cmark & \textit{Temper.} & \xmark & \xmark & \xmark & \xmark & \xmark & EN \\
3MDBench~\citep{sviridov2025mdbench}       & Benchmark & Med images      & 3k    & \cmark & \cmark & \textit{Temper.} & \xmark & \xmark & \xmark & \xmark & \cmark & EN \\
MedAgentSim~\citep{almansoori2025medagentsim} & Benchmark & Mixed       & 637   & \cmark & \cmark & \xmark           & \xmark & \xmark & \xmark & \xmark & \xmark & EN \\
\midrule
\rowcolor{gray!15}
\textbf{MedRoundsQA (Ours)} & Benchmark & \textbf{Board exams} & 1.4k &
\cmark & \cmark & \textbf{Multi-attr} &
\textbf{\cmark} & \textbf{\cmark (24)} &
\textbf{\cmark} & \textbf{\cmark} & \cmark & EN \\
\bottomrule
\end{tabular}
}
\caption{\textbf{Comparison with existing medical QA and dialogue benchmarks.}
\textbf{MT}: multi-turn dialogue.
\textbf{MA}: multi-agent (doctor--patient) setting.
\textbf{Persona}: systematic variation of patient attributes (\textit{Temper.}=emotion/temperament; \textit{Multi-attr.}=multiple controlled attributes such as education, affect, language proficiency, occupation).
\textbf{DF}: automatic difficulty stratification.
\textbf{Schema}: canonical structured clinical representation (\textit{(24)} = 24-slot record).
\textbf{CF (Counterfactual personas)}: same underlying case instantiated across personas (enables causal attribution of persona effects).
\textbf{ST$\leftrightarrow$MT}: single-turn and multi-turn evaluated on the same normalized cases under the same ground truth, so the ST--MT gap isolates the interactive component; benchmarks reporting both settings without a shared canonical representation are not marked.
\textbf{FC}: full consultation until diagnostic decision or turn budget.
\textbf{Key novelty:} the only benchmark combining a canonical schema, counterfactual personas, and same-case ST$\leftrightarrow$MT evaluation, enabling controlled measurement of communication-driven reliability/fairness gaps.}
\label{tab:comparison}
\end{table*}

We introduce \emph{MedRoundsQA}, converting 1{,}387 board-exam questions across 17 specialties into controlled doctor--patient dialogues. Each case is normalized into a \textbf{24-slot clinical record} serving as ground truth, then instantiated with \textbf{counterfactual patient personas} varying communication characteristics (health literacy, language proficiency, affect, occupation) while holding clinical content fixed. Cases are further stratified by difficulty via model-based uncertainty signals.

Our evaluation of fifteen LLM doctor agents yields three findings: (i) diagnostic accuracy drops 13--39 points moving from single-turn to multi-turn on the same cases; (ii) additional turns improve question relevance but diagnostic accuracy plateaus after 6--12 turns; (iii) patient personas shift accuracy by 7--8 points across education levels, exposing equity concerns invisible to single-turn evaluation.

\paragraph{Contributions.}
\textbf{(1) Counterfactual causal control:} each case is instantiated
under multiple personas with clinical content held fixed in a shared
24-slot record, attributing performance gaps to communication style
rather than case difficulty and enabling same-case
ST$\leftrightarrow$MT comparison.
(2) \textbf{Difficulty-aware evaluation:} A model-uncertainty protocol stratifies cases for fine-grained easy-to-hard analysis.
(3) \textbf{Empirical characterization of interaction failures:} Substantial ST$\rightarrow$MT degradation, diminishing returns with longer dialogues, and systematic persona-driven disparities.

\section{Related Work}
\textsc{MedRoundsQA} evaluates consultation-style diagnosis where both multi-turn interaction and patient presentation matter. Measuring the effect of patient communication requires counterfactual control: instantiating multiple personas for the same underlying case to avoid confounding by case difficulty.

\paragraph{Medical QA benchmarks.}
Exam-style benchmarks MedQA~\citep{jin2020medqa}, MedMCQA~\citep{pal2022medmcqa}, PubMedQA~\citep{jin2019pubmedqa}, BioASQ~\citep{tsatsaronis2015bioasq}, MultiMedQA~\citep{singhal2023largelanguage}, and MMLU~\citep{hendrycks2021mmlu} assess factual recall but treat each case as a single-shot vignette, ignoring incremental information acquisition inherent in real consultations.

\paragraph{Medical dialogue corpora.}
Large-scale corpora (MedDialog~\citep{he2020meddialog}, MedDG~\citep{liu2022meddg}, IMCS-21~\citep{chen2022imcs21}, ChatDoctor~\citep{li2023chatdoctor}) support training but are opportunistic collections with heterogeneous quality and weak ground-truth labels, unsuited for controlled evaluation.

\paragraph{Multi-agent medical dialogue benchmarks.}
AgentClinic~\citep{schmidgall2024agentclinic}, MediQ~\citep{li2024mediq}, AI Hospital~\citep{fan2024aihospital}, Dr.APP~\citep{zhu2025drapp}, 3MDBench~\citep{sviridov2025mdbench}, and MedAgentSim~\citep{almansoori2025medagentsim} introduce agent-based frameworks, some with persona modeling. Unlike these, \textsc{MedRoundsQA} uses a canonical evidence interface with same-case persona instantiation, enabling counterfactual evaluation of communication-driven performance gaps.

\paragraph{Multi-turn medical consultation evaluation.}
Most closely related work lacks the controls needed to isolate patient communication effects.
\citet{liao2309automatic} use a single fixed patient prompt with no persona variation, making it impossible to attribute performance differences to communication style rather than case difficulty.
MedDialogRubrics~\citep{gong2026meddialogrubrics} measures process quality against EBM-derived rubrics, but its ground truth is LLM-generated and cases are never held fixed across patient styles, precluding causal persona attribution.
\citet{manczak2025shallow} evaluate adversarial robustness after a diagnosis is formed from a complete vignette; there is no patient agent and no information elicitation, making it orthogonal to consultation-style reasoning.
MedMT-Bench~\citep{yang2025medmtbench} targets instruction-following and memory across long contexts rather than diagnostic accuracy, with no validated ground truth or persona control.
\textsc{MedRoundsQA} is the only benchmark combining validated board-exam ground truth, a 24-slot canonical schema, same-case counterfactual personas, difficulty stratification, and direct ST$\leftrightarrow$MT comparison on identical clinical content- capabilities absent from all four prior works (Table~\ref{tab:comparison}).

\section{Methods}
We construct a controlled benchmark that converts single-turn exam questions into grounded multi-turn doctor--patient consultations with (i) a canonical 24-slot case record, (ii) systematically varied patient personas, and (iii) difficulty labels derived from model-based uncertainty.

To convert vignettes into our 24-slot schema, we compare three candidate LLMs (GPT-4o~\cite{openai-gpt4o-system-card-2024}, DeepSeek~\cite{deepseek-api-docs-deepseek-chat}, and Claude~\cite{anthropic-claude-4-system-card-2025}) on a randomly sampled set of 100 questions. A general practitioner (GP) manually annotates these cases into the schema to provide slot-level reference records. We evaluate extraction quality \emph{slot-wise} (micro-averaged across slots) using lexical and semantic similarity measures and select the model that performs best overall. GPT-4o consistently achieves the strongest agreement with GP annotations (Fig.~\ref{fig:model_selection_bars}), and we use it to extract structured records for the full benchmark.

\subsection{Data Construction}
\label{sec:data-construction}

\paragraph{Source and filtering.} We derive cases from MedQA~\citep{jin2020medqa}, NEJM diagnostic cases~\citep{savage2024diagnostic}, and MedMCQA~\citep{pal2022medmcqa}. We filter to retain only text-sufficient diagnostic vignettes, excluding items focused on management/guidelines/factual recall or those requiring images or non-clinical administrative knowledge. Filtering uses a two-stage protocol: an automatic LLM filter (Appendix~\ref{appendix:filter-prompt}) followed by independent review by two GPs; disputed inclusions are removed. The final dataset contains 1{,}387 cases (292 MedQA, 646 NEJM, 449 MedMCQA) spanning 17 specialties.

% \paragraph{Normalizing questions and answers.}
% Because sources are multiple-choice, we remove answer options from the model input and retain only the vignette and question text. We map each answer key to its option text and normalize it into a canonical \texttt{ground\_truth\_diagnosis} string used for evaluation.

\paragraph{Canonical case representation.}
To bridge heterogeneous question styles and support dialogue generation,
we convert each vignette into a unified structured representation that
decouples clinical content from its surface realization. Our schema is
inspired by the structured medical records used in MMD-Eval \citep{liu2025interactive}, but adapted to outpatient diagnostic vignettes
and extended to 24 clinical entities plus a final diagnosis field. The schema
is organized into four groups mentioned in Appendix \ref{appendix:clinical_entities_groups}.

We implement a two-stage annotation procedure to balance scale and clinical faithfulness. First, we use \texttt{GPT-4o}
to extract information and map each vignette into the schema. The model reads
the vignette and populates as many slots as are explicitly supported by the
text, copying short phrases rather than paraphrasing (e.g., ``right upper
quadrant pain'' or ``temperature is 38.3$^\circ$C''). Missing information is
marked as \textsc{NA} rather than hallucinated. Second, to ensure dataset
correctness and high quality, three resident physicians manually revise
the converted records. We partition the dataset across the three annotators,
and each annotator compares the extracted slot values against the original
vignette, correcting typos and formatting, adding any clinically relevant
information missed by \texttt{GPT-4o}, and removing any unsupported or
hallucinated content. In an additional normalization pass, we enforce
cross-case consistency (e.g., units, spelling, and slot boundaries) and schema
constraints such as non-empty \texttt{Symptom-Name} for all records (annotator instructions in Appendix~\ref{app:annotator_instructions}). This step
also assigns each case to one of the 17 specialties used in our analyses. The
full schema template, including clinical entity names and descriptions, is
given in Appendix~\ref{appendix:standard-format}. The resulting structured
records serve as the ground truth for both dialogue generation and evaluation,
and are the only source of clinical truth used throughout our benchmark. Examples of full records are provided in Appendix~\ref{appendix:example-standard-format}.

\subsection{Automatic Difficulty Estimation}

We assign each case a difficulty level using a model-based proxy grounded in prior findings that (i) longer reasoning traces correlate with harder problems and (ii) model uncertainty/disagreement can reflect item difficulty~\citep{wu2025more,zotos2025can}. For each question $q_i$, we run an ensemble of $K{=}4$ reasoning LLMs with a shared structured prompt and sample each model $N{=}6$ times. From these runs, we compute three signals: (1) average chain-of-thought length $L_i$ (number of steps), (2) semantic answer disagreement $H_i$ estimated via accuracy over the $N$ final answers per LLM, and (3) a confidence proxy $C_i$ that combines answer frequency and self-reported confidence within exact-match answer clusters.

We min-max normalize $(L_i, H_i, C_i)$ across questions and define a continuous difficulty score
\begin{equation}
d_i = \frac{\tilde L_i + \tilde H_i + (1-\tilde C_i)}{3}.
\end{equation}
Finally, we cluster $\{d_i\}$ with 1D $k$-means ($k{=}3$) and map clusters by increasing mean $d_i$ to \emph{Easy}, \emph{Medium}, and \emph{Hard}. Full model/prompt details are in Appendix~\ref{app:difficulty}; the
labels are further validated against blinded physician annotations,
reaching 88.9\% exact agreement with physician majority labels
(weighted $\kappa = 0.88$; Appendix~\ref{app:difficulty-validation}).

\subsection{Selecting the Patient Simulator}
To isolate \emph{doctor}-side effects in our evaluations, we fix a single patient-simulator model for all experiments. We compare three candidates (GPT-4o, Claude, DeepSeek-Chat) using a controlled record-grounding test: for each candidate, we generate 100 dialogues against a fixed doctor and score each doctor question as \textsc{Hit}/\textsc{Miss} under three rules (answerable $\rightarrow$ note-consistent answer; unanswerable $\rightarrow$ explicit ``don't know''; symptom absent from the note $\rightarrow$ ``No''). We repeat this comparison under three fixed doctor models (DeepSeek-Chat, GPT-4o, Claude). DeepSeek-Chat achieves the highest or near-highest adherence across settings, so we use DeepSeek-Chat as the patient simulator in all subsequent experiments. See Appendix~\ref{app:patient_selection} for full results. To verify that results are not driven by the simulator choice, we fix the doctor to GPT-4o and vary only the patient model on 100 stratified cases. Diagnosis accuracy is similar across simulators (within 2 pp), indicating low sensitivity to the patient model under our record-grounded protocol (Appendix~\ref{app:patient_selection}).

\subsection{Patient Personas}
To isolate the effect of patient communication on diagnostic performance, we generate controlled patient personas that vary only in how the same clinical facts are expressed. Personas span four axes: education/health literacy, emotional tone, language proficiency, and occupational framing. For each clinical case, all personas share an identical underlying structured note; only linguistic style and interaction behavior change. This enables controlled analysis of robustness and fairness (e.g., whether doctor models perform differently for low-literacy or non-native patients). Full persona definitions and examples are provided in Appendix~\ref{appendix:dialogue-creation-process}.

\subsection{Dialogue Generation Pipeline}
\label{sec:dialogue-generation}

Given a canonical structured record and a patient persona, we synthesize multi-turn doctor-patient consultations using two LLM agents (doctor and patient) coordinated by a lightweight session controller. Figure~\ref{fig:dialoug} provides an overview of the pipeline. Details and prompts are in Appendix \ref{appendix:dialogue-creation-process}.

\paragraph{Doctor agent.}
The doctor agent is prompted to conduct a realistic history-taking interview by asking focused follow-up questions, while avoiding treatment recommendations and not revealing access to the underlying record. To preserve the interactive setting, the doctor is given only a brief chart blurb derived from the case (e.g., exam/notes), rather than the full structured record.

\paragraph{Patient agent.}
The patient agent is prompted with (i) a persona description and (ii) a redacted case report constructed from the structured record. The \texttt{ground\_truth\_diagnosis} field is withheld to prevent label leakage. The patient answers strictly based on this report and explicitly indicates uncertainty (e.g., ``I am not sure'') when the doctor asks about missing information.

% \paragraph{Session controller.}
% A session controller alternates turns between the doctor and patient agents and maintains their conversation histories until termination.

\paragraph{Termination and provisional diagnosis.}
Each consultation runs for up to a maximum number of turns. The doctor may diagnose earlier or continue until the limit. At termination, the doctor outputs a single-line provisional diagnosis; we store this line and the full transcript for evaluation.

\subsection{Evaluation Methods}

\paragraph{Single-turn baseline on original questions.}
To contextualize multi-turn performance, we first evaluate each model on the original datasets in a single-turn setting. We remove multiple-choice options and prompt the model with only the vignette and question, requiring a free-form diagnosis. Because diagnoses may have non-identical surface forms, we use an automatic LLM judge to map each response to a binary correct/incorrect label (Appendix~\ref{appendix:single_turn_eval_prompt}) and report accuracy per model.

\begin{table*}[t!]
\centering
\small
\setlength{\tabcolsep}{2.2pt}
\renewcommand{\arraystretch}{0.92}
\resizebox{\linewidth}{!}{
    \begin{tabular}{l r r r r r r r}
    \toprule
    Model & Turns & Dx(ST)~(\%) & Dx(MT)~(\%) & \cellcolor{gray!15}\textbf{Drop}$\downarrow$ & SlotCov~(\%) & RelAcc~(\%) & EarlyTerm~(\%) \\
    \midrule
    \multicolumn{8}{l}{\textbf{Closed-source}} \\
    GPT-4o~\citep{openai-gpt4o-system-card-2024}                        & 13.57 & 67.47 {\small±2.02} & 31.55 {\small±2.09} & \cellcolor{gray!15}\textbf{35.92} & 10.20 & 97.26 & 90.70 \\
    DeepSeek-Chat~\citep{deepseek-api-docs-deepseek-chat}               & 16.68 & 66.10 {\small±2.05} & 27.14 {\small±1.98} & \cellcolor{gray!15}\textbf{38.96} & 26.99 & 93.93 & 71.46 \\
    Gemini-2.5-Flash~\citep{comanici2025gemini}                         & 15.92 & 65.75 {\small±2.13} & 26.68 {\small±1.95} & \cellcolor{gray!15}\textbf{39.07} & 25.74 & 96.40 & 69.76 \\
    Claude Sonnet 4.5~\citep{anthropic-claude-sonnet-45-system-card-2025} & 18.30 & 63.50 {\small±2.10} & 24.66 {\small±1.95} & \cellcolor{gray!15}\textbf{38.84} & 30.00 & 97.96 & 44.86 \\
    
    Seed-2.0-Mini~\citep{bytedanceseed2026seed2}                                                       & 10.95 & 52.40 {\small±2.22} & 25.56 {\small±1.94} & \cellcolor{gray!15}\textbf{26.84} & 20.39 & 92.12 & 55.08 \\
    \midrule
    \multicolumn{8}{l}{\textbf{Open-source}} \\
    Llama3-UltraMed~\citep{zhang2024ultramedical}                       & 15.59 & 50.00 {\small±2.20} & 22.21 {\small±1.84} & \cellcolor{gray!15}\textbf{27.79} & 19.81 & 89.79 & 63.03 \\
    HuatuoGPT~\citep{chen2025huatuogpt}                                      & 10.16 & 60.62 {\small±2.16} & 35.85 {\small±2.13} & \cellcolor{gray!15}\textbf{24.77} & 37.23 & 89.95 & 96.40 \\
    Qwen3-VL-32B-Instruct~\citep{qwen3-vl-technical-report-2025}       & 16.20 & 63.00 {\small±2.06} & 28.77 {\small±2.03} & \cellcolor{gray!15}\textbf{34.23} & 18.50 & 98.50 & 99.70 \\
    Baichuan-M2-32B~\citep{baichuanm2-scaling-medical-capability-2025}  & 17.60 & 48.00 {\small±2.18} & 19.86 {\small±1.90} & \cellcolor{gray!15}\textbf{28.14} & 22.00 & 96.88 & 90.10 \\
    Llama3-OpenBioLLM-8B~\citep{aaditya-llama3-openbiollm-8b-2024}     & 15.50 & 45.00 {\small±2.24} & 16.78 {\small±1.88} & \cellcolor{gray!15}\textbf{28.22} & 20.00 & 96.68 & 69.70 \\
    Phi-4~\citep{abdin2024phi4}                                                               & 11.33 & 33.22 {\small±2.10} & 19.78 {\small±1.77} & \cellcolor{gray!15}\textbf{13.44} & 23.96 & 93.24 & 84.97 \\
    Ministral3-8B~\citep{mistral2026ministral3}                                                        & 10.68 & 36.30 {\small±2.14} & 19.92 {\small±1.78} & \cellcolor{gray!15}\textbf{16.38} & 19.93 & 92.38 & 72.08 \\

    Kimi-K2.5~\citep{kimi-k25-visual-agentic-intelligence-2026}         & 13.73 & 72.00 {\small±1.98} & 41.92 {\small±2.12} & \cellcolor{gray!15}\textbf{30.08} & 14.00 & 97.93 & 88.01 \\
    MiniMax-M2.5~\citep{minimax2026m25}                                                        & 10.77 & 60.27 {\small±2.18} & 36.40 {\small±2.14} & \cellcolor{gray!15}\textbf{23.87} & 11.60 & 97.23 & 89.44 \\
    GLM-5~\citep{zeng2026glm}                                                                & 14.09 & 64.73 {\small±2.13} & 31.21 {\small±2.06} & \cellcolor{gray!15}\textbf{33.52} & 28.19 & 97.44 & 66.92 \\
    \bottomrule
    \end{tabular}
}
\caption{\textbf{Main results on \textsc{MedRoundsQA}. Values are percentages except Turns.} Dx(ST) is single-turn accuracy using the standardized record-style input; Dx(MT) is multi-turn diagnostic accuracy. \textbf{Drop} = Dx(ST) $-$ Dx(MT) (higher indicates a larger degradation under multi-turn interaction).}
\label{tab:main-results}
\end{table*}

\paragraph{Single-turn from structured records.}
To verify that our canonical schema preserves diagnostic signal and to measure robustness to input format, we also evaluate models on the structured record (omitting ground\_truth\_diagnosis). Models again output a free-form diagnosis, which is scored with the same judge protocol. Comparing this accuracy to the vignette-based baseline quantifies information retention and sensitivity to the narrative-to-structured shift.

\paragraph{Locked validation of LLM judges.}
Since we rely on an LLM judge for diagnosis correctness and other dialogue-level metrics, we validate it with a locked protocol. We stratified-sampled 300 dialogues and obtained independent labels from three resident physicians; majority vote serves as the reference and we report inter-rater reliability and judge agreement. We finalized the judging rubric/prompt on a 100-dialogue development split and then froze it before evaluation on a held-out 200-dialogue split to mitigate prompt overfitting. On the held-out split, GPT-4o achieved the highest agreement with the human majority-vote labels for diagnosis correctness (approximately 0.88--0.90 accuracy; $\kappa \approx 0.70$--0.80), outperforming alternative judge models. Finally, all structured case records used for dialogue generation are human-verified.

% \section{Experiments}
% Hardware specs
% Inference settings (temperature, top-p, max tokens)
% Number of runs / seeds for variance
% Why these specific models (GPT-4o, Gemini, DeepSeek, HuatuoGPT, UltraMedical)?

\section{Experiment Results}

Table~\ref{tab:main-results} summarizes multi-turn performance on \textsc{MedRoundsQA} under a 20-turn budget. Across models, interactive diagnosis is roughly half of single-turn performance on the standardized record: a drop of $13$--$39$ points. Question relevance stays high ($90$--$99\%$), so models mostly stay on topic, but they gather little structured information (slot coverage $10$--$37\%$) and stop early in most cases ($45$--$100\%$ of dialogues). Behavioral patterns do not translate into better interactive diagnosis. Kimi-K2.5, for example, asks relatively few questions (low slot coverage, $14\%$) yet achieves the best multi-turn accuracy ($42\%$). HuatuoGPT covers the most slots ($37\%$) and achieves strong multi-turn accuracy ($36\%$), but still terminates early $96\%$ of the time. MiniMax-M2.5 uses even fewer turns than Kimi-K2.5 (avg.\ 10.77) and achieves a moderate 36\% accuracy despite one of the lowest slot coverages of any model (11.6\%). Simply asking more questions or filling more fields rarely helps; models struggle to identify and act on the right discriminative findings.

\paragraph{Evaluation metrics.}
Let $y_i$ be the gold diagnosis for case $i$, and let $\hat{y}_i^{ST}$, $\hat{y}_i^{MT}$ denote the single-turn and multi-turn predictions. \textbf{Dx(ST/MT)} $= \frac{1}{N}\sum_{i=1}^{N}\mathbb{1}[\hat{y}_i^{(\cdot)}=y_i]$ measures diagnosis accuracy under each setting; \textbf{Drop} $= \mathrm{Dx(ST)}-\mathrm{Dx(MT)}$ quantifies degradation under interaction. Let $S_i$ be the set of non-empty gold schema slots for case $i$, and let $e_{i,s}=\mathbb{1}[\exists\,t: \operatorname{elicits\_slot}(q_{i,t},a_{i,t})=s]$; then \textbf{SlotCov} $= \frac{1}{N}\sum_i \frac{1}{|S_i|}\sum_{s\in S_i} e_{i,s}$ measures the fraction of available clinical information elicited. \textbf{RelAcc} $= \frac{1}{N}\sum_i \frac{1}{T_i}\sum_{t=1}^{T_i} r_{i,t}$ is the fraction of doctor questions judged relevant or grounded in the record by an LLM judge, and \textbf{EarlyTerm} $= \frac{1}{N}\sum_i \mathbb{1}[T_i < T_{\max}]$ is the rate of stopping before the turn budget. Full formal definitions and judge prompts are in Appendices~\ref{app:metrics} and~\ref{appendix:evaluation-prompts}.

\begin{figure}[h]
  \centering
  \includegraphics[width=\linewidth]{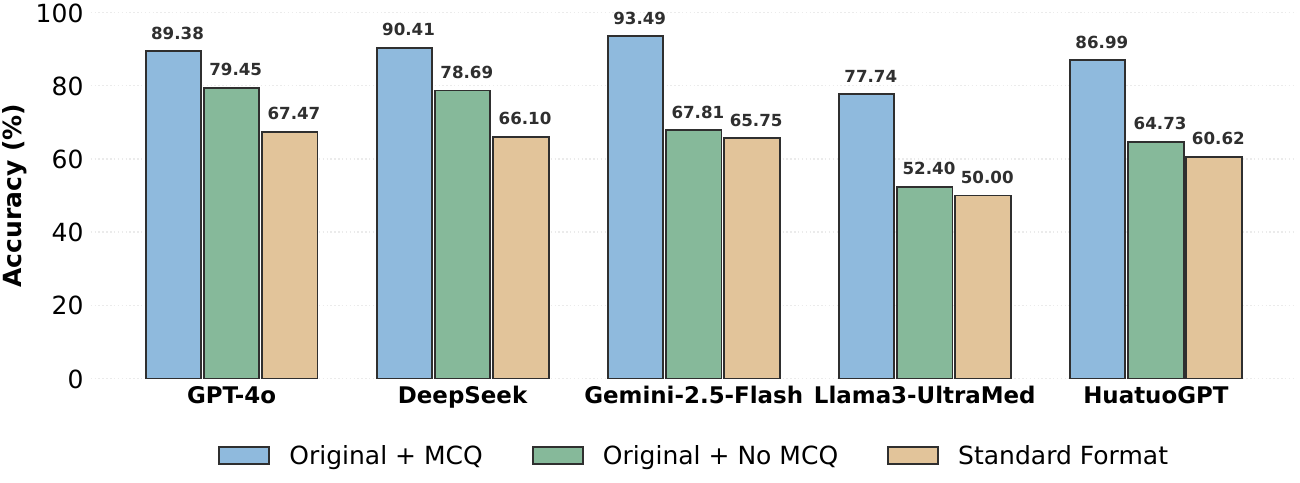}
  \caption{\textbf{Effect of format on diagnostic accuracy.} We compare the multiple-choice setting (\emph{+MCQ}), the same cases without answer options (\emph{No MCQ}), and a standardized record-style representation (\emph{Standard Format}).}

  \label{fig:medqa-format-accuracy}
  \vspace{-1.0em}
\end{figure}

\subsection{Impact of Question Format}
To disentangle multi-turn effects from input-format scaffolding, we evaluate models under three settings when supported by the source data: \emph{+MCQ} (with answer options), \emph{No MCQ} (open-ended diagnosis), and a standardized record-style \emph{Standard Format} (Figure~\ref{fig:medqa-format-accuracy}).
Across models, most of the “accuracy” comes from exploiting answer choices rather than doing genuine open-ended reasoning: stripping MCQs cuts performance by 10-26 points while further standardizing the input only costs a few additional points, echoing findings that LLMs often solve MedQA-style benchmarks via option elimination and pattern matching rather than full case understanding (e.g., on MedQA and MedMCQA). This means multi-turn results should be compared to the open-ended, standardized settings, not MCQ scores, which substantially overstate a model’s diagnostic ability in realistic, non-MCQ clinical workflows.

\begin{figure*}[t!]
  \centering
  \includegraphics[width=0.92\textwidth, trim={15 5 15 5}, clip]{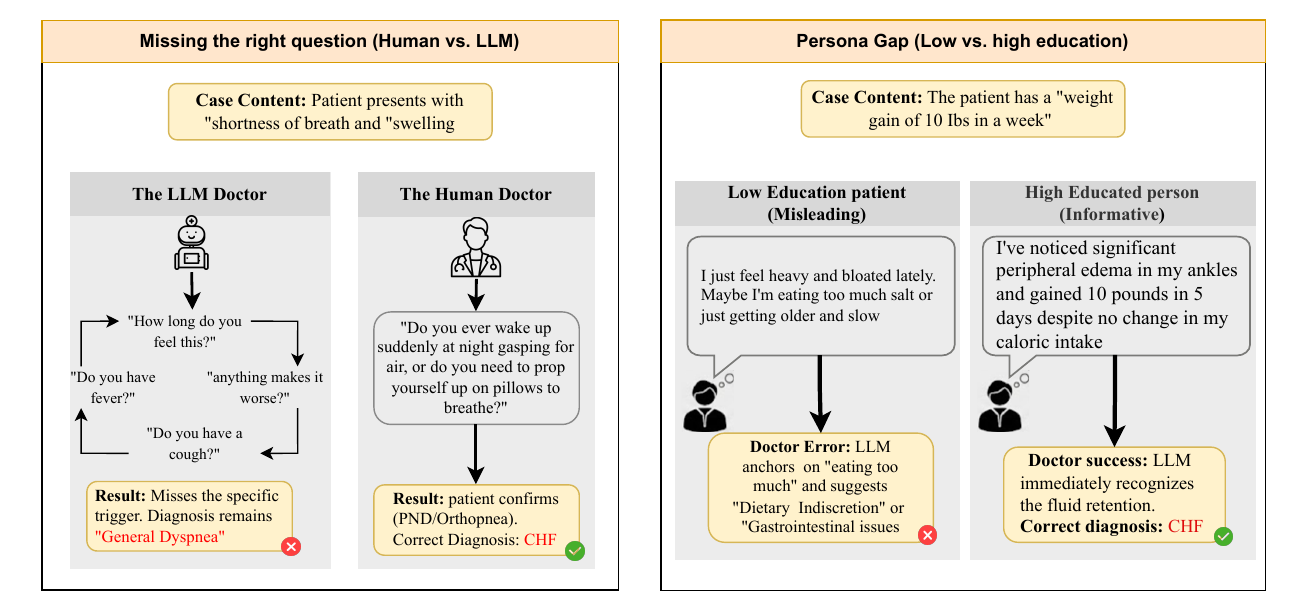}
  \caption{\textbf{Persona-driven elicitation failure.} Holding clinical content constant, an informative persona foregrounds high-yield details that trigger targeted questioning and correct diagnosis (CHF), while a misleading persona yields vague framing that leads to generic questioning/anchoring and an incorrect or underspecified diagnosis.}
  \label{fig:qualitative-example}
  \vspace{-1.0em}
\end{figure*}

\subsection{Can LLMs Match Human Performance?}
\label{sec:LLMvsHuman}
\begin{table}[h!]
\centering
\rowcolors{2}{white}{gray!10}
\resizebox{\columnwidth}{!}{%
\begin{tabular}{lcccc}
\toprule
\textbf{Model/Physician} & \textbf{Correct (n/60)} & \textbf{Acc (\%)} & \textbf{Avg.\ turns} & \textbf{Min.\ turns} \\
\midrule
\multicolumn{5}{c}{\texttt{AI models}} \\
GPT-4o                  & 17/60 & 28.33 & 11.63 & 6 \\
DeepSeek                & 15/60 & 25.00 & 16.00 & 8 \\
Gemini-2.5-Flash        & 19/60 & 31.67 & 15.17 & 8 \\
Llama3-UltraMedical-70B & 14/60 & 23.33 & 13.55 & 6 \\
HuatuoGPT               & 15/60 & 25.00 & 5.27 & 4 \\
\textbf{AI average}     & 16.0/60 & 26.67 & 12.32 & 6.4 \\
\midrule
\multicolumn{5}{c}{\texttt{Human physicians}} \\
Physician 1             & 29/60 & 48.33 & 7.69 & 2 \\
Physician 2             & 31/60 & 51.67 & 8.50 & 6 \\
Physician 3             & 27/60 & 45.00 & 5.59 & 2 \\
\textbf{Human average}  & 29/60 & 48.33 & 7.26 & 3.3 \\
\midrule
\textbf{Perf.\ gap (Human - AI)} 
                        & +13.0 & +21.66 p & -5.06 & -3.1 \\
\bottomrule
\end{tabular}%
}
\caption{Human vs.\ LLM diagnostic performance on 60 challenging cases using a shared simulated patient.}

\label{tab:human_model_performance}
\end{table}

To ground our interactive evaluation in a human clinical baseline, we compare LLM “doctor’’ agents to resident physicians using the same 20-turn simulated-patient interface on 60 difficult cases (Table~\ref{tab:human_model_performance}), and find that resident physicians nearly double LLM diagnostic accuracy ($48\%$ vs.\ $27\%$) while using $40\%$ fewer turns, echoing prior findings that clinicians still outperform frontier models on non-MCQ, case-based diagnosis despite strong results on static benchmarks (e.g., Med-PaLM and GPT-4 evaluations). The most “efficient” model actually underperforms, stopping after ~5 turns with only $25\%$ accuracy, highlighting that current LLMs tend to terminate prematurely rather than strategically probing for high-yield clinical evidence. Qualitatively, human doctors reliably ask targeted, discriminative questions (e.g., orthopnea/paroxysmal nocturnal dyspnea in Figure~\ref{fig:qualitative-example}), whereas LLMs default to generic screening queries and are easily steered off-course by low-literacy or misleading personas, leading to under-specified or incorrect diagnoses even when key clues are present.

Our findings align with recent clinical evidence: in a randomized
controlled study with 1,298 participants, \citet{bean2026reliability}
found that LLMs identified conditions in 94.9\% of medical scenarios
when tested alone, yet participants assisted by the same models
succeeded in fewer than 34.5\% of cases--a collapse under
interaction consistent with the ST$\to$MT drops we observe, and one
that motivates controlled interactive benchmarks as a scalable
evaluation layer before costly human studies.

\begin{table}[h]
    \centering
    \footnotesize
    \resizebox{\linewidth}{!}{
        \rowcolors{2}{white}{gray!10}
        \begin{tabular}{lccc}
            \toprule
            Model & Easy & Medium & Hard \\
            \midrule
            GPT-4o                   & 38.68 & 22.46 & 10.07 \\
            Gemini-2.5-Flash         & 43.15 & 25.31 & 14.99 \\
            DeepSeek-Chat            & 37.93 & 23.39 & 12.49 \\
            Claude Sonnet 4.5        & 32.10 & 18.45 &  8.21 \\
            Kimi-K2.5                & 54.20 & 31.85 & 18.43 \\
            MiniMax-M2.5             & 47.30 & 27.60 & 14.82 \\
            GLM-5                    & 40.55 & 23.89 & 11.34 \\
            Seed-2.0-Mini            & 33.25 & 18.82 &  8.97 \\
            Llama3-UltraMedical      & 30.10 & 17.79 &  5.20 \\
            HuatuoGPT                & 28.10 & 16.11 &  5.31 \\
            Qwen3-VL-32B-Instruct    & 37.20 & 21.43 &  9.88 \\
            Baichuan-M2-32B          & 25.90 & 14.72 &  5.01 \\
            Llama3-OpenBioLLM-8B     & 22.15 & 12.38 &  4.19 \\
            Phi-4                    & 26.40 & 14.95 &  5.88 \\
            Ministral3-8B            & 26.85 & 15.21 &  5.62 \\
            \bottomrule
        \end{tabular}
    }
    \caption{Diagnostic accuracy (\%) stratified by difficulty band
             (Easy/Medium/Hard), using DeepSeek-Chat as the fixed
             patient simulator with a 12-turn budget.}
    \label{tab:difficulty-bands}
    \vspace{-1.5em}
\end{table}
\subsection{Performance Across Difficulty Bands}
\label{sec:difficulty-results}

Table~\ref{tab:difficulty-bands} stratifies diagnostic accuracy by Easy/Medium/Hard cases using our model-based difficulty scores, with DeepSeek as the fixed patient simulator and a 12-turn budget. The difficulty signal transfers cleanly to the interactive setting: for every model, accuracy declines monotonically as cases get harder.
Qualitatively, Easy cases tend to be ones where a single discriminative clue (e.g. classic risk-factor constellation, a pathognomonic exam finding) is present and, once elicited, most models converge on the same diagnosis within a few turns. Hard cases, by contrast, often require integrating multiple weak signals or reconciling competing explanations (e.g., overlapping cardiopulmonary or rheumatologic presentations); in these settings, we observe the behaviors highlighted in our error taxonomy: models hedge (“could be X or Y’’), default to syndrome-level labels, or terminate without ever asking the one or two questions that human annotators considered decisive.

\begin{figure}[h]
    \vspace{-0.5em}
    \centering
    \includegraphics[width=\linewidth]{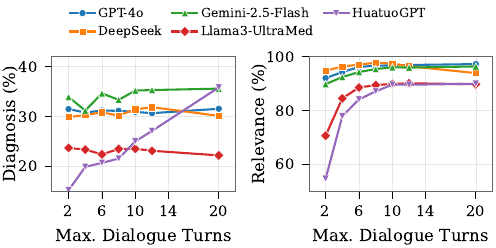}
    \vspace{-1.5em}
    \caption{Diagnosis accuracy and question relevance as a function of maximum dialogue turns.}
    \label{fig:turn-budget}
    \vspace{-1.5em}
\end{figure}

\subsection{How Much Dialogue Is Enough?}
\label{sec:turn_budget}
Varying the maximum consultation length shows that models quickly learn to stay on-topic: question relevance rises steeply with more turns and then saturates, yet diagnostic accuracy barely improves beyond a short interview (Fig.~\ref{fig:turn-budget}). In most cases, additional turns produce more of the same low-yield questions rather than sharper hypotheses or better synthesis of gathered clues (e.g. Figure~\ref{fig:qualitative-example}); only HuatuoGPT meaningfully benefits from longer budgets, suggesting that the central bottleneck is not talking longer but turning incremental history into a decisive diagnostic update.

\begin{figure}[t]
  \centering
  \includegraphics[width=\linewidth]{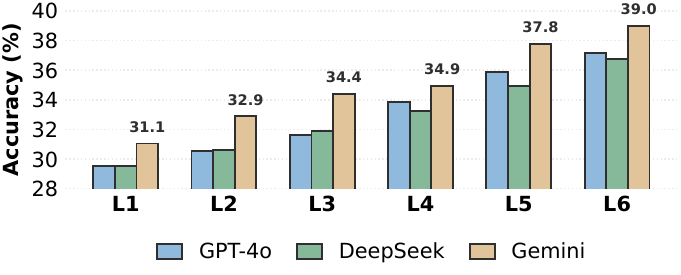}
  \vspace{-1.5em}
  \caption{Diagnostic accuracy by education (health-literacy) persona. Accuracy increases monotonically with education level across all models.}
  \label{fig:persona-education}
  \vspace{-1.0em}
\end{figure}

\begin{figure}[t]
  \centering
  \includegraphics[width=\linewidth]{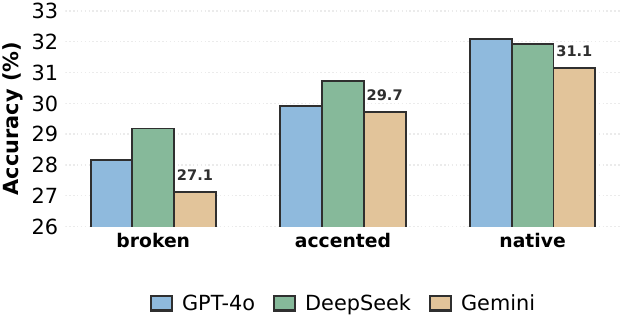}
  \caption{Diagnostic accuracy by language-proficiency persona. Lower fluency (broken language) yields substantial performance drops.}
  \label{fig:persona-language}
\end{figure}

% \subsection{Persona Analysis}
% \label{sec:persona_analysis}

% We probe robustness to \emph{how} patients communicate by varying education, language fluency, occupation, and affect while holding clinical content fixed. Education (health literacy) is the strongest driver: all models diagnose markedly better as histories become more organized and medically precise, producing a near-monotonic rise from the lowest to highest education levels (Figure~\ref{fig:persona-education}). Lower language proficiency and blue-collar occupational framing further depress accuracy, whereas affect has only modest, model-dependent effects, suggesting that surface clarity matters more than emotional tone. For example, the persona comparison in Figure~\ref{fig:qualitative-example} shows how a low-literacy patient who talks about “eating too much salt’’ leads the model to generic gastrointestinal labels, while a high-literacy version of the same case foregrounds rapid weight gain and edema and reliably elicits the key orthopnea/PND question, underscoring that many errors arise from failures to extract and integrate high-yield information from noisy narratives. Figure \ref{fig:persona-language} shows that the better the language the better is the accuracy. See more detailed results in Appendix~\ref{appendix:case-dialogues}.

\subsection{Persona Analysis}
\label{sec:persona_analysis}

We probe robustness to how patients communicate by varying education, language fluency (including accent), occupation, and affect while holding clinical content fixed. Education (health literacy) is the strongest driver: all models diagnose markedly better as histories become more organized and medically precise, producing a near-monotonic rise from the lowest to highest education levels (Figure~\ref{fig:persona-education}). Reduced language proficiency and, to a lesser extent, accented or non-native phrasing further depress accuracy, and blue-collar occupational framing also lowers performance, whereas affect has only modest, model-dependent effects, suggesting that surface clarity matters more than emotional tone. For example, the persona comparison in Figure~\ref{fig:qualitative-example} shows how a low-literacy patient who talks about “eating too much salt’’ leads the model to generic gastrointestinal labels, while a high-literacy version of the same case foregrounds rapid weight gain and edema and reliably elicits the key orthopnea/PND question, underscoring that many errors arise from failures to extract and integrate high-yield information from noisy narratives. Figure~\ref{fig:persona-language} shows that stronger language skills are associated with higher accuracy. See more detailed results in Appendix~\ref{app:acc_per_persona}.

\begin{table}[t!]
    \resizebox{\linewidth}{!}{
        \rowcolors{2}{white}{gray!10}
        \begin{tabular}{lcccc}
            \toprule
            \textbf{Error Type} & \textbf{Overall} & \textbf{GPT-4o} & \textbf{DeepSeek} & \textbf{Gemini} \\
            \midrule
            Protocol breakdown       & 9\%  & 6\%  & 11\% & 8\%  \\
            Uncertainty-as-output    & 40\% & 35\% & 48\% & 42\% \\
            Syndrome-level collapse  & 13\% & 10\% & 15\% & 14\% \\
            Insufficient elicitation & 30\% & 28\% & 35\% & 30\% \\
            \bottomrule
        \end{tabular}
    }
    \caption{\textbf{Distribution of error types across models.} Percentages are the fraction of incorrect predictions (943 cases) exhibiting each pattern (multi-labeling allowed). Models differ primarily in frequency rather than the presence/absence of distinct error types.}
    \label{tab:error-types}
\end{table}
\section{Why LLMs Fail Interactive Diagnosis?}
\label{sec:LLMs_fail_reasons}

We manually annotate a sample of 943 incorrect multi-turn consultations drawn from the GPT-4o, DeepSeek-Chat, and Gemini-2.5-Flash runs, using a shared taxonomy, finding four recurring failure modes
(Table~\ref{tab:error-types}). \textbf{Protocol breakdowns}
(9\%) occur when the model never produces a usable diagnosis.
\textbf{Uncertainty-as-output} (40\%), the most prevalent
mode, occurs when models hedge or retreat into generic workup
language even after sufficient evidence has been elicited.
\textbf{Insufficient elicitation} (30\%) captures cases where
models never ask the few high-yield questions human reviewers
considered decisive (e.g., orthopnea/PND for heart failure,
specific exposures for infectious presentations, family history
for hereditary conditions, or orthostatic vitals for volume status). \textbf{Syndrome-level collapse} (13\%) occurs when models settle on broad labels despite the
case supporting a more specific entity. Crucially, error-type distributions are near-identical across GPT-4o, DeepSeek, and Gemini---stronger models fall into them less often but not less \emph{consistently}---indicating that scaling has not resolved the core weaknesses: knowing when to commit, identifying discriminative questions, and synthesizing partial histories into granular diagnoses. These patterns are invisible
to single-turn evaluation, where the same models score
50--67\% on standardized records (Table~\ref{tab:main-results}),
but emerge clearly under interactive conditions. The low-literacy
persona analysis (Section~\ref{sec:persona_analysis}) further shows that
communication noise amplifies all four failure modes, with
anchoring and insufficient elicitation rising sharply when
patients use vague or misleading symptom descriptions. The
clinician audit (Appendix~\ref{app:clinician-audit}) confirms:
successful dialogues show structured history-taking and differential-aware consolidation (61\%, 48\%), while failed ones exhibit repetition and anchoring (57\%, 49\%), pointing to a
shared bottleneck in strategic clinical reasoning. Representative error cases with clinical commentary are in Appendix~\ref{appendix:case-dialogues}.

\section{Conclusion}

MedRoundsQA converts exam vignettes into controlled multi-turn consultations, allowing direct comparison of single-turn and interactive diagnosis on identical cases. Across fifteen LLMs, multi-turn evaluation reveals weaknesses missed by single-turn scores: accuracy falls sharply from records to consultations, plateaus after about 6–12 turns despite more relevant questions, and varies systematically with patient communication features including education, language fluency, and occupational framing. In matched simulations, resident physicians nearly double model accuracy using fewer turns, while error analysis shows models often stop early, hedge rather than commit, and miss key discriminative findings. Improving medical LLMs requires stronger knowledge plus better targeted questioning and diverse communication styles.

\newpage
\section{Limitations and Future Work}

We focus on text-based, single-encounter consultations and do not model longitudinal records or imaging; these are natural extensions for future work. Our results suggest that moving beyond single-shot vignettes to controlled multi-turn settings is both feasible and necessary if we want to understand how LLMs behave in the kinds of consultations clinicians and patients are beginning to conduct with AI tools today.

\paragraph{Protocol design and setup-induced effects.}
Three protocol choices could in principle shape the observed
failures. The brief chart blurb mirrors the information state of a
new-patient encounter \citep{schmidgall2024agentclinic}; since Dx(ST)
(Table~\ref{tab:main-results}) measures the same cases under full information access, the
ST$\to$MT gap isolates the interactive component. The turn budget is
not binding: accuracy plateaus well before 20 turns
(Section~\ref{sec:turn_budget}). And under identical prompts and decoding,
early-termination rates still vary from 44.9\% to 99.7\% (Table~\ref{tab:main-results}),
while resident physicians on the same interface nearly doubled model
accuracy with fewer turns (Section~\ref{sec:LLMvsHuman}). The observed weaknesses are
thus primarily model-intrinsic, though alternative blurb contents,
adaptive budgets, and prompt variations merit future study.

Beyond protocol design, while the consistency of failure-mode distributions across
our 15 models spanning 8B open medical models to frontier
systems suggests a bottleneck not resolved by scale alone
(Section~\ref{sec:LLMs_fail_reasons}), a controlled within-family scaling analysis, comparing
models of different sizes from the same family under identical
protocols, would provide a cleaner test of this hypothesis and is a
concrete direction for future work.

\section{Ethical Statement}
We introduce \textsc{MedRoundsQA}, a benchmark for evaluating medical reasoning in multi-turn simulated patient--clinician dialogues built from publicly available, de-identified cases converted into structured records and synthetic conversations. The benchmark does not involve collecting new patient data, interacting with real patients, or conducting any human-subjects intervention.

Because the setting is medical, outputs or examples could be misconstrued as clinical advice. Our dataset and evaluation are intended solely for research on model behavior and robustness and are not a substitute for professional diagnosis or decision-making. We also evaluate controlled persona variations (e.g., language proficiency, education, affect), which may reveal performance disparities; we avoid targeting protected attributes and report these effects to encourage safer and more equitable medical dialogue systems.

The data used in this study were obtained exclusively from publicly available sources. No new data were collected directly from individuals. The original data were released under terms that permit research use. Our work involves creating a derived dataset through processing and annotation of this public data. As such, informed consent was handled at the time of the original data collection by the respective data providers, and no additional consent was required for this study.

Large language models were used in a limited capacity to assist with minor editing and polishing of the manuscript, such as improving clarity and grammar. All technical content, experimental design, results, and conclusions were produced, verified, and finalized by the authors.

The physician annotators were recruited through the authors' academic and
clinical networks on the basis of their medical training and relevant
clinical expertise. General practitioners and resident physicians
participated in dataset filtering, structured-record verification,
difficulty-label validation, and dialogue-quality evaluation. All annotators
provided informed consent and were compensated at a fixed rate of
5~USD per hour, which substantially exceeds the typical effective hourly
earnings of resident physicians in Egypt. No crowdsourcing platforms or
members of the general public were involved.
\bibliography{custom}

\appendix
\clearpage
\appendix

% Optional: Reset section numbering style for appendix
\renewcommand{\thesection}{\Alph{section}}
\renewcommand{\thesubsection}{\thesection.\arabic{subsection}}

% Add a title page for appendix (optional but nice)
\section*{Appendix}
\raggedbottom
\addcontentsline{toc}{section}{Appendices}

This appendix provides implementation details, prompts, and supplementary materials for \textsc{MedRoundsQA}.

% ==============================================================================
\section{Data Collection and Filtering}
\label{appendix:data-collection}
% ==============================================================================

\subsection{Filtering Prompt}
\label{appendix:filter-prompt}

We use the following prompt to automatically filter medical questions for suitability:

\begin{promptbox}
\small
\texttt{You are helping to filter a dataset of medical multiple-choice questions.}

\texttt{Your task is to decide whether the question is a *clinical vignette that asks for a diagnosis* and provides enough information about the patient to make a diagnosis.}

\texttt{A question qualifies if:}\\
\texttt{- It presents a clinical scenario or patient vignette, and}\\
\texttt{- The main goal is to identify the most likely diagnosis for the patient.}

\texttt{If the question meets both criteria, respond with: YES}\\
\texttt{Otherwise, respond with: NO}

\texttt{Respond with a single word only: YES or NO.}

\texttt{QUESTION:}\\
\texttt{"question text"}
\end{promptbox}

\section{Dataset Construction and Extraction Details}
\label{appendix:extraction_and_filtering}

\subsection{GP reference annotations for extractor selection}
We recruit a general practitioner (GP) to convert 100 development questions into the 24-slot schema, extracting only information explicitly stated in the vignette and marking missing information as \texttt{NA}. This produces slot-level reference records used to select an automatic extractor.

\subsection{Extractor evaluation protocol and metrics}
For each candidate extractor (GPT-4o, DeepSeek, Claude), we generate a structured record for each of the 100 development questions using the same slot definitions provided to the GP. We score outputs in a \emph{slot-aware} manner: each predicted slot value is compared to the corresponding GP-annotated slot value and scores are micro-averaged across all slots, penalizing slot misplacements.

\subsection{Data Extractor Selection}

\begin{figure}[h!]
  \centering
  \includegraphics[width=\linewidth]{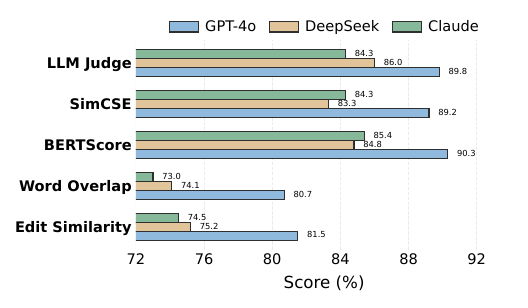}
  \caption{Comparison of extraction models across five evaluation metrics against GP-annotated ground truth.}
  \label{fig:model_selection_bars}
\end{figure}

We report both lexical and semantic similarity metrics, including normalized Levenshtein similarity, Jaccard word overlap, BERTScore-F1 (implemented via \texttt{bert-score}), and cosine similarity of supervised SimCSE embeddings (\texttt{sup-simcse-bert-base-uncased}). We also include an LLM-as-a-judge adequacy score using \texttt{claude-sonnet-4-20250514} with a 0.90 threshold. Results are shown in the Figure \ref{fig:model_selection_bars}.

% \subsection{Filtering prompt and adjudication}
% We first apply an automatic LLM filter with a fixed binary prompt (Appendix~\ref{appendix:filter-prompt}) to identify text-sufficient diagnostic vignettes. Two independent GPs then re-label all items accepted by the filter using the same criterion; items with disagreement are excluded to avoid ambiguous inclusions. We provide the full prompt, examples of exclusions, and additional breakdowns in this section.

\subsection{Answer normalization}
We map each MCQ answer key to its corresponding option text and normalize it into a canonical \texttt{ground\_truth\_diagnosis} (e.g., resolving letter indices and standardizing minor surface-form variation).

\section{Metrics and Formal Definitions}
\label{app:metrics}

We formally define all metrics reported in Table~\ref{tab:main-results}. Let the dataset be indexed as $\mathcal{D}=\{1,\dots,N\}$. For each case $i\in\mathcal{D}$, let the gold diagnosis be $y_i$, and let $\hat{y}_i^{ST}$ and $\hat{y}_i^{MT}$ denote the single-turn and multi-turn predictions, respectively. Let the multi-turn dialogue have length $T_i$ with questions $q_{i,t}$ and answers $a_{i,t}$ for $t\in\{1,\dots,T_i\}$. Let $T_{\max}$ be the maximum turn budget, and let $S_i$ be the set of gold non-empty slots for case $i$. Let $\mathbb{1}[P]=1$ if predicate $P$ holds and $0$ otherwise.

\paragraph{Dx(ST).}
\begin{equation}
\mathrm{Dx(ST)}=\frac{1}{N}\sum_{i=1}^{N}\mathbb{1}\!\left[\hat{y}_i^{ST}=y_i\right].
\end{equation}

\paragraph{Dx(MT).}
\begin{equation}
\mathrm{Dx(MT)}=\frac{1}{N}\sum_{i=1}^{N}\mathbb{1}\!\left[\hat{y}_i^{MT}=y_i\right].
\end{equation}

\paragraph{Drop.}
\begin{equation}
\mathrm{Drop}=\mathrm{Dx(ST)}-\mathrm{Dx(MT)}.
\end{equation}

\paragraph{SlotCov.}
Define
\begin{equation}
e_{i,s}=\mathbb{1}\!\left[\exists\, t\le T_i:\ \operatorname{elicits\_slot}(q_{i,t},a_{i,t})=s\right].
\end{equation}
Then
\begin{equation}
\begin{aligned}
\mathrm{SlotCov}(i)=\frac{1}{|S_i|}\sum_{s\in S_i} e_{i,s}.
\end{aligned}
\end{equation}
\begin{equation}
\mathrm{SlotCov}=\frac{1}{N}\sum_{i=1}^{N}\mathrm{SlotCov}(i).
\end{equation}

\paragraph{RelAcc.}
Define
\begin{equation}
r_{i,t}=\mathbb{1}\!\left[q_{i,t}\ \text{is relevant/grounded}\right],
\end{equation}
where relevance/grounding is determined by an LLM judge with a fixed rubric.
Then
\begin{equation}
\mathrm{RelAcc}(i)=\frac{1}{T_i}\sum_{t=1}^{T_i} r_{i,t}.
\end{equation}
\begin{equation}
\mathrm{RelAcc}=\frac{1}{N}\sum_{i=1}^{N}\mathrm{RelAcc}(i).
\end{equation}

\paragraph{EarlyTerm.}
\begin{equation}
\mathrm{EarlyTerm}(i)=\mathbb{1}\!\left[T_i<T_{\max}\right].
\end{equation}
\begin{equation}
\mathrm{EarlyTerm}=\frac{1}{N}\sum_{i=1}^{N}\mathrm{EarlyTerm}(i).
\end{equation}

\section{Dataset Statistics}
\label{appendix:datasets_statistics}
\begin{figure*}[t]
  \centering
  \includegraphics[width=\linewidth]{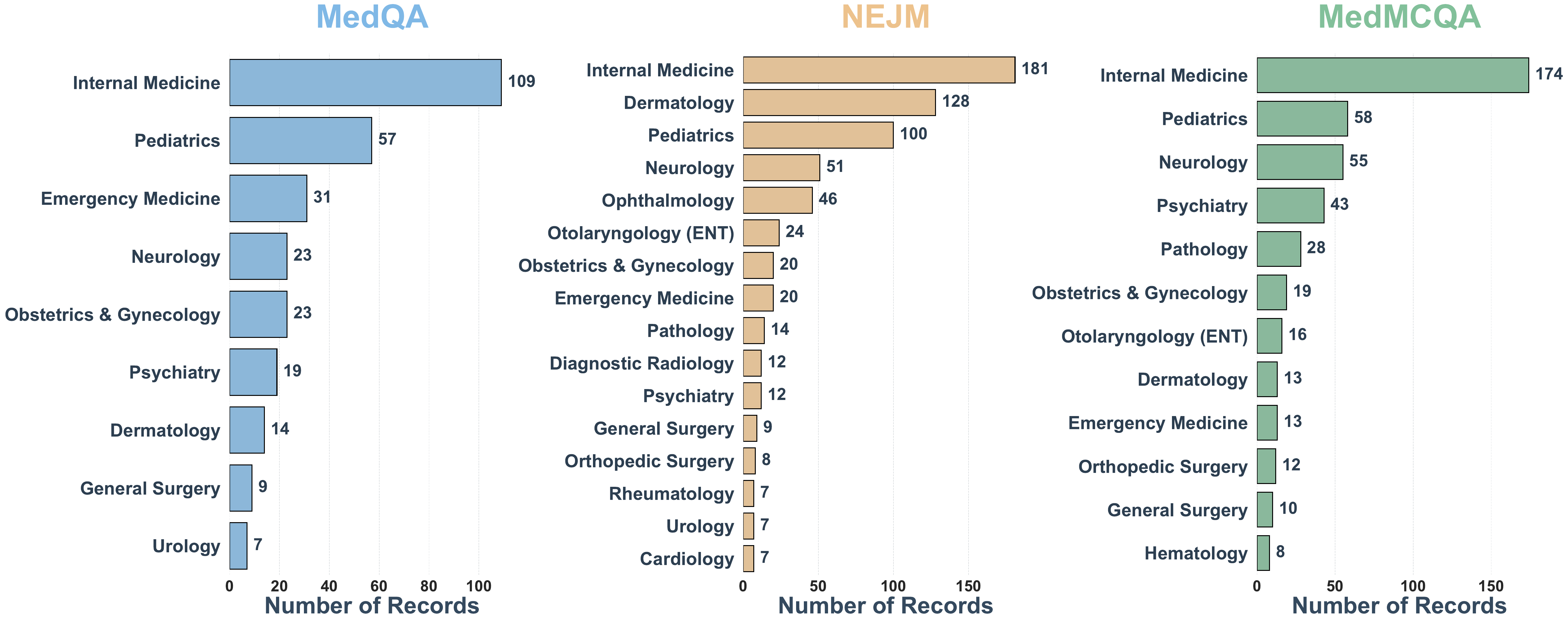}
  \caption{Specialty distribution for each dataset subset.}
  \label{fig:specialty_distribution}
\end{figure*}

Figure~\ref{fig:specialty_distribution} shows the distribution of medical
specialties in each subset of the dataset. Internal Medicine is the largest
specialty in all three sources, and NEJM contributes the broadest specialty
coverage (16 of the 17 specialties), while several specialties (e.g.,
Rheumatology, Cardiology) contain fewer than ten cases each consistent
with the small-sample caution for such specialties in
Appendix~\ref{sec:specialty-results}.

\section{Clinical Record Schema}
\label{appendix:standard-format}
% ==============================================================================

\subsection{Schema Definition}

Each clinical case is represented using a structured 24-slot schema organized into four semantic groups. Table~\ref{tab:schema-full} provides complete slot definitions.

\begin{table}[]
\centering
\scriptsize
\setlength{\tabcolsep}{3pt}
\begin{tabularx}{\linewidth}{p{3.2cm}X}
\toprule
\textbf{Slot Name} & \textbf{Definition} \\
\midrule
\multicolumn{2}{l}{\textit{\textbf{Symptom Descriptors}}} \\
\texttt{Symptom-Name} & Primary symptoms reported by the patient in lay (non-medical) terms. Avoid disease names; describe only what a non-professional would notice. \\
\texttt{Symptom-Time} & Onset time, duration, and course of the symptoms (e.g., acute, chronic, intermittent). \\
\texttt{Symptom-Position} & Anatomical location of the symptoms. \\
\texttt{Symptom-Degree} & Severity or intensity of the symptoms (e.g., mild, moderate, severe). \\
\texttt{Symptom-Color} & Color features of the symptoms (e.g., redness, cyanosis, jaundice). \\
\texttt{Symptom-Shape} & Shape or form of visible findings (e.g., patches, nodules, swelling). \\
\texttt{Symptom-Texture} & Qualitative characteristics of the symptoms (e.g., thickness of phlegm, firmness of a lump). \\
\texttt{Symptom-Smell} & Odor related to the symptoms (e.g., foul-smelling discharge, fruity breath). \\
\midrule
\multicolumn{2}{l}{\textit{\textbf{Physical Changes \& Examination}}} \\
\texttt{Physical-Change} & Noticeable changes in physical state such as weight, appetite, energy level, or sleep pattern. \\
\texttt{Medical-Examination} & Findings from physical examination and additional tests (e.g., CT, MRI, X-ray, laboratory tests, ECG). \\
\midrule
\multicolumn{2}{l}{\textit{\textbf{History \& Context}}} \\
\texttt{Personal-Age} & Patient's age. \\
\texttt{Personal-Sex} & Patient's sex/gender as recorded in the medical context. \\
\texttt{Personal-Weight} & Patient's body weight (and height if available). \\
\texttt{Past-Diagnosis} & Record of previous medical diagnoses, including major past illnesses. \\
\texttt{Past-Medication} & Medications the patient is currently taking or has recently taken on a regular basis. \\
\texttt{Trauma-Surgery} & History of significant trauma, injuries, or surgical procedures. \\
\texttt{Allergic-History} & History of allergies to foods, medications, or other substances, and typical reactions. \\
\texttt{Preventive} & Preventive measures and vaccination status, including recent or missing vaccines. \\
\texttt{Family-History} & Family medical history, including major illnesses, hereditary or genetic diseases. \\
\texttt{Contact-History} & Recent contact with infectious diseases, sick individuals, animals, or high-risk environments. \\
\texttt{Habit-Tobacco} & Tobacco use, including type, amount, and duration. \\
\texttt{Habit-Wine} & Alcohol use (wine and other alcoholic beverages), including frequency and amount. \\
\texttt{Habit-Drug} & Use or abuse of illicit drugs or non-prescribed psychoactive substances. \\
\texttt{Habit-Living} & Lifestyle habits such as diet, exercise, hygiene, work pattern, and sleep routine. \\
\texttt{Coitus-History} & History of sexual activity, especially unprotected intercourse or multiple partners when relevant. \\
\texttt{Receiving-Treatment} & Whether the patient is currently receiving treatment and a brief description of that treatment. \\
\midrule
\multicolumn{2}{l}{\textit{\textbf{Meta-Information}}} \\
\texttt{Department} & Clinical department or specialty where the patient is being evaluated (e.g., cardiology, pediatrics). \\
\texttt{additional\_notes} & Any other relevant patient information or medication details not covered by the above categories. \\
\texttt{ground\_truth\_diagnosis} & Definitive diagnosis provided by a qualified medical professional after evaluation. \\
\bottomrule
\end{tabularx}
\caption{\textbf{Complete schema specification.} Each slot captures a specific aspect of the clinical presentation, history, or examination findings.}
\label{tab:schema-full}
\end{table}

\subsection{Clinical entities groups}
\label{appendix:clinical_entities_groups}

The 24 slots are organized into four semantic groups. Table~\ref{tab:schema} summarizes each group, its member slots, and its role in representing a case: symptom descriptors capture the presenting complaint and its attributes, physical changes and examination findings encode objective measurements, history and context provide background risk factors, and meta-information stores the specialty label and the canonical diagnosis used for evaluation.

\begin{table}[h]
\centering
\small
\resizebox{\linewidth}{!}{
\begin{tabular}{p{2.8cm} p{4.5cm} p{5.2cm}}
\toprule
\textbf{Group} & \textbf{Example slots} & \textbf{Role in case representation} \\
\midrule
Symptom descriptors 
& \texttt{Symptom-Name}, \texttt{Symptom-Time}, \texttt{Symptom-Position}, \texttt{Symptom-Degree}, \texttt{Symptom-Color}, \texttt{Symptom-Shape}, \texttt{Symptom-Texture}
& Capture the core complaint and its attributes (what the patient feels, when it started, where it is, and how severe or distinctive it is). \\
\addlinespace[0.3em]
Physical changes and examination findings 
& \texttt{Physical-Change}, \texttt{Medical-Examination}
& Encode measurable findings such as vital signs, focused physical examination, and key laboratory or imaging results. \\
\addlinespace[0.3em]
History and context 
& \texttt{Personal-Age}, \texttt{Personal-Sex}, \texttt{Personal-Weight}, \texttt{Past-Diagnosis}, \texttt{Past-Medication}, \texttt{Trauma-Surgery}, \texttt{Allergic-History}, \texttt{Preventive}, \texttt{Family-History}, \texttt{Contact-History}, \texttt{Habit-Tobacco}, \texttt{Habit-Wine}, \texttt{Habit-Drug}, \texttt{Habit-Living}, \texttt{Coitus-History}
& Provide background risk factors, lifestyle factors, and longitudinal information that modulate prior probability of different diagnoses. \\
\addlinespace[0.3em]
Meta-information 
& \texttt{Department}, \texttt{additional\_notes}, \texttt{ground\_truth\_diagnosis}
& Store specialty classification, brief free-text clarifications, and the canonical diagnosis used as the evaluation label. \\
\bottomrule
\end{tabular}
}
\caption{Overview of the 24-slot clinical schema used to represent each clinical case. Clinical entities are grouped into four categories that separate presenting symptoms, objective findings, history, and meta-information.}
\label{tab:schema}
\end{table}

\subsection{Frequency of Clinical Entity Groups}
\begin{figure}[h]
    \centering
    \includegraphics[width=\linewidth]{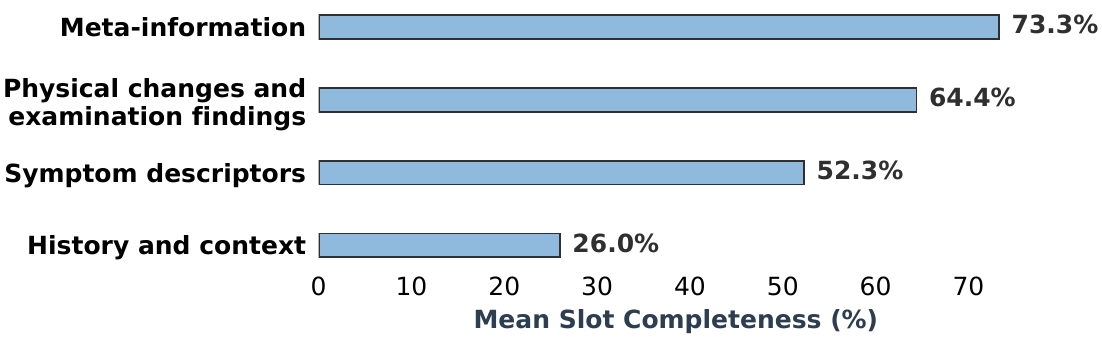}
    \caption{Mean slot completeness (\% non-\textsc{NA}) aggregated by schema group.}
    \label{fig:mean_slot_pct_by_group}
\end{figure}
We further quantify annotation coverage by measuring, for each slot, the
percentage of cases with a non-\textsc{NA} value, and then averaging these
slot-level completeness scores within each group. Figure~\ref{fig:mean_slot_pct_by_group}
summarizes the resulting mean slot completeness per group. We note that this
distribution is inherently imbalanced: not every slot is clinically informative
for every vignette, and our schema records only information that is supported
by the original question and potentially useful for reaching the correct
diagnosis. As a result, groups such as history and context may appear sparser,
even when the underlying cases are well-specified for diagnosis.

\subsection{Example of a record in a standard format}
\label{appendix:example-standard-format}
Below is a complete example of a case represented in our standardized format:

\lstdefinestyle{jsonstyle}{
  basicstyle=\small\ttfamily,
  columns=fullflexible,
  keepspaces=true,
  showstringspaces=false,
  breaklines=true,
  breakatwhitespace=true
}

\begin{tcolorbox}[
  breakable,
  enhanced,
  colback=gray!5,
  colframe=black!70,
  boxrule=0.5pt,
  arc=2pt,
  left=8pt,right=8pt,top=8pt,bottom=8pt
]
\begin{lstlisting}[style=jsonstyle]
{
  "record": {
    "Symptom-Name": "increasing shortness of breath, dry cough",
    "Past-Diagnosis": "NA",
    "Past-Medication": [],
    "Physical-Change": "difficulty walking, puffy and red face,
      pitting edema of the legs",
    "Trauma-Surgery": "NA",
    "Preventive": "NA",
    "Allergic-History": "NA",
    "Contact-History": "NA",
    "Habit-Tobacco": "smokes 1.5 packs of cigarettes a day",
    "Habit-Wine": "drinks about four bottles of beer a day",
    "Habit-Drug": "NA",
    "Habit-Living": "NA",
    "Coitus-History": "NA",
    "Family-History": "NA",
    "Personal-Age": "55",
    "Personal-Weight": "NA",
    "Personal-Sex": "male",
    "Medical-Examination": "chest X-ray discloses hyperinflation,
      flattening of the diaphragm, increased retrosternal air space",
    "Department": "NA",
    "Receiving-Treatment": "NA",
    "Symptom-Degree": "breathless after only a few steps",
    "Symptom-Position": "chest, legs (pitting edema)",
    "Symptom-Time": "past few years",
    "Symptom-Shape": "barrel chest",
    "Symptom-Texture": "NA",
    "Symptom-Smell": "NA",
    "Symptom-Color": "NA",
    "additional_notes": "Prolonged expiration with wheezing,
      hyperresonance on percussion, clubbing of the digits",
    "ground_truth_diagnosis": "Emphysema"
  }
}
\end{lstlisting}
\end{tcolorbox}

\section{Data difficulty classification}
\label{app:difficulty}

\paragraph{Models and prompting.}
We use an ensemble of $K{=}4$ reasoning LLMs: \textsc{DeepSeek-R1}, \textsc{Qwen/QwQ-32B}, \textsc{Phi-4-reasoning}, and \textsc{OLMo-3.1-32B-Think}. Each model is queried with the same structured prompt (Appendix~\ref{app:diff_prompt}) and sampled $N{=}6$ times per question.

\paragraph{Signals.}
For each question $q_i$ and model $M_k$, we collect $N$ runs producing a reasoning trace, final answer, and self-reported confidence $c\in[0,1]$. We compute:
(i) mean CoT length $L_{ik}$ (steps),
(ii) semantic disagreement $H_{ik}\in[0,1]$ by sending the set of $N$ final answers to a judge model,
and (iii) a cluster-based confidence proxy. We canonicalize answers (lowercasing, whitespace normalization), form exact-match clusters, and for each cluster $a$ compute frequency $f_a$ and mean confidence $\bar c_a$. We define $S_a=\tfrac12(f_a+\bar c_a)$ and set $C_{ik}$ to the mean $S_a$ over runs. We average across models to obtain $L_i,H_i,C_i$.

\paragraph{Difficulty score and binning.}
We min--max normalize $L_i,H_i,C_i$ across all questions to get $\tilde L_i,\tilde H_i,\tilde C_i$ and compute
\begin{equation}
d_i = \frac{\tilde L_i + \tilde H_i + (1-\tilde C_i)}{3}.
\end{equation}
We then apply 1D $k$-means ($k{=}3$) to $\{d_i\}$ and label clusters as \emph{Easy}/\emph{Medium}/\emph{Hard} by increasing mean $d_i$.

\subsection{Difficulty Prompt}
\label{app:diff_prompt}
\begin{promptbox}
\small
You are a medical reasoning assistant. Be concise and factual.
YOU MUST OUTPUT EXACTLY ONE JSON OBJECT ON A SINGLE LINE AND NOTHING ELSE.
Do not print any extra commentary, markdown, or code fences.
For the question below, return EXACTLY one JSON object with keys:
\texttt{"cot"} (array of numbered step strings), \texttt{"final\_answer"} (1--3 words), and
\texttt{"confidence"} (number 0--1). Do NOT output anything else.

\par\medskip
\textbf{Question:}\\
\texttt{\{question\_text\}}

\par\medskip
\textbf{Example (single-line JSON exactly like this):}\\
\texttt{\{"cot":["1. Step one","2. Step two"],"final\_answer":"Histoplasma capsulatum","confidence":0.87\}}
\end{promptbox}

\subsection{Data Difficulty distribution}
\begin{figure}[H]
  \centering
  \includegraphics[width=\linewidth]{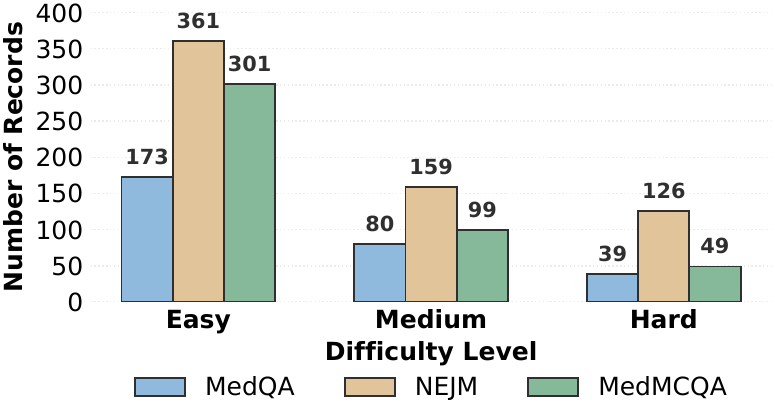}
  \caption{Distribution of difficulty levels across dataset samples.}
  \label{fig:dataset_difficulty}
\end{figure}

\subsection{Physician Validation of Difficulty Labels}
\label{app:difficulty-validation}
To verify that our model-based difficulty labels correspond to
clinically meaningful difficulty, we conducted a blinded physician
annotation study on 90 cases sampled evenly across the
Easy/Medium/Hard bands. Physicians rated difficulty from the
original vignettes without access to the automatic labels or any
model outputs. The automatic labels achieved 88.9\% exact agreement
with the physician majority labels, with weighted Cohen's
$\kappa = 0.88$ (linear weights) and $\kappa = 0.92$ (quadratic
weights). All disagreements occurred between adjacent bands, with
no Easy--Hard confusions (Table~\ref{tab:difficulty-confusion}).

\begin{table}[h]
\centering
\begin{tabular}{lccc}
\toprule
 & \multicolumn{3}{c}{Automatic label} \\
\cmidrule(lr){2-4}
Physician label & Easy & Medium & Hard \\
\midrule
Easy   & 27 & 3  & 0  \\
Medium & 2  & 25 & 3  \\
Hard   & 0  & 2  & 28 \\
\bottomrule
\end{tabular}
\caption{Confusion matrix between automatic difficulty labels and
blinded physician majority labels on 90 stratified cases.}
\label{tab:difficulty-confusion}
\end{table}

\section{Annotator Instructions}
% \subsection{Information Extraction Annotation Instructions}
\label{app:annotator_instructions}
\begin{promptbox}
\small
You will be shown an original clinical vignette together with a structured
clinical record automatically extracted from that vignette. Your task is to
verify that the structured record accurately reflects the information
explicitly stated in the original vignette.
\par\medskip
\textbf{Please follow these rules:}
\begin{enumerate}[leftmargin=*, itemsep=1pt, topsep=2pt]
  \item Review each structured field against the original vignette.
  \item Correct any factual errors, typos, formatting problems, or
        information placed in the wrong field.
  \item Add clinically relevant information that is explicitly stated in
        the vignette but missing from the structured record.
  \item Remove any information that is not supported by the vignette.
  \item Do not infer, guess, or add clinical information that is not
        explicitly stated in the vignette.
  \item If information for a field is unavailable, retain or assign the
        value \texttt{NA}.
  \item Preserve the original clinical meaning and, where possible, use
        short phrases taken directly from the vignette rather than
        introducing new interpretations.
  \item Verify that the final diagnosis corresponds to the diagnosis
        provided by the original source.
\end{enumerate}
\par\medskip
If a case is ambiguous, or you are uncertain whether a piece of information
is supported by the vignette, flag the case for review rather than making an
unsupported assumption.
\par\medskip
The purpose of this task is to produce a clinically faithful structured
representation of each case. These records are intended for research
evaluation only and should not be interpreted as real-patient medical
records or clinical advice.
\end{promptbox}

\section{Patient Simulator Selection and Robustness}
\label{app:patient_selection}

\begin{figure}[H]
  \centering
  \includegraphics[width=\linewidth]{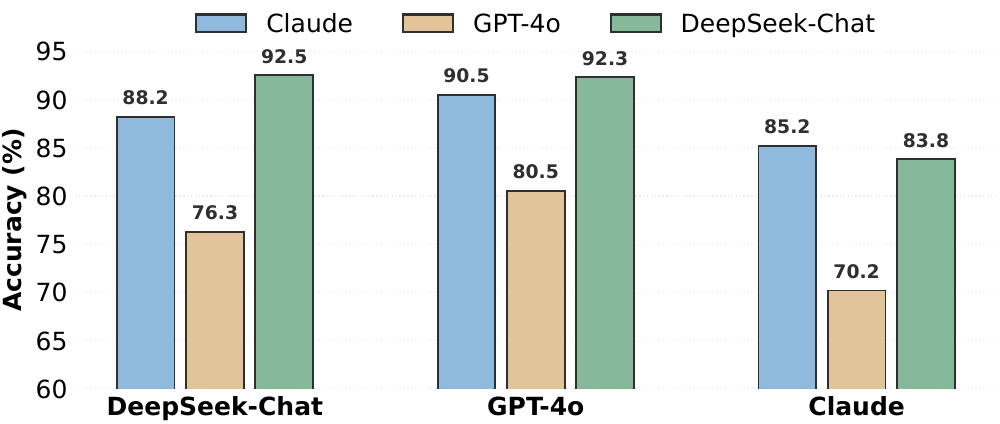}
  \caption{Comparison of patient models under the record-grounded adherence test.}
  \label{fig:patient_model_selection}
\end{figure}

\paragraph{Selection protocol.}
We evaluate GPT-4o, Claude, and DeepSeek-Chat as patient simulators by generating 100 dialogues (600 turns) per candidate against a fixed doctor model. Each doctor question is scored as \textsc{Hit} if it follows record-grounding rules: (i) if answerable from the structured note, the patient response must be consistent with the note; (ii) if not answerable, the patient must reply with an explicit ``don't know''; (iii) if the doctor asks about a symptom absent from the note, the patient must answer ``No''. Accuracy is the fraction of turns scored as \textsc{Hit}. We repeat the evaluation with three fixed doctor models (DeepSeek-Chat, GPT-4o, Claude). DeepSeek-Chat yields the highest or near-highest adherence across all three doctor settings (Fig.~\ref{fig:patient_model_selection}).

\paragraph{Robustness check.}
We fix the doctor to GPT-4o and vary only the patient simulator (DeepSeek-Chat, GPT-4o, Claude) using identical prompts/decoding and a redacted patient record that excludes \texttt{ground\_truth\_diagnosis}. On a stratified subset of 100 cases (25 easy/50 medium/25 hard), final diagnosis accuracy is 30.0\% (DeepSeek-Chat), 31.0\% (GPT-4o), and 29.0\% (Claude). Paired bootstrap confidence intervals show simulator-induced differences are small (e.g., $\Delta(\text{DeepSeek}-\text{GPT-4o})=-1.0$pp; $\Delta(\text{DeepSeek}-\text{Claude})=+1.0$pp).

\section{Dialogue Generation System Prompts}
\label{appendix:dialogue-creation-process}
% ==============================================================================

\subsection{Doctor Agent Prompt}

The doctor agent is instructed to conduct a focused medical history using standard clinical heuristics.

\begin{tcolorbox}[
  breakable,
  enhanced,
  colback=gray!5,
  colframe=black!70,
  boxrule=0.5pt,
  arc=2pt,
  left=8pt,right=8pt,top=8pt,bottom=8pt
]
You are a kind, careful family doctor speaking in clear, friendly, everyday language.

\textbf{Your role:}
\begin{itemize}
\item Begin the conversation yourself.
\item Just start the conversation by saying ``Hello, how can I help you today?''
\item It is enough to ask them why they are here today.
\item You have not seen the patient before—act as if you know nothing about their case.
\item You have access to a brief `chart blurb' with objective info (e.g., vitals, exam notes). Mimic that you tested those vitals/exams during your questioning and got the answer.
\item MIMIC like you are a real doctor who is taking temperature for the patient to see if they have fever or not, or checking their blood pressure.
\item Ask focused, empathetic, clarifying questions to gather key medical history.
\end{itemize}

\textbf{Guidelines for questioning:}
\begin{itemize}
\item Explore symptoms using OPQRST (onset, provocation, quality, radiation, severity, timing).
\item Screen for red flags, pregnancy, medications, allergies, past medical/surgical history, and relevant social habits.
\item Ask about personal and family history when relevant (e.g., smoking, alcohol, drug use, chronic illnesses, age, sex, weight, pregnancy status, sexual activity).
\item Ask one topic at a time—do not bundle unrelated questions.
\item Be warm, empathetic, and conversational.
\end{itemize}

\textbf{Chart Blurb:} \texttt{[Medical-Examination field + additional notes]}

\textbf{Restrictions:}
\begin{itemize}
\item Do NOT prescribe or recommend specific medications or treatments.
\item Do NOT mention or imply that you can see the full patient report—only refer to the visible chart blurb.
\item The patient will answer based STRICTLY on a PRIVATE structured medical report you do NOT see.
\end{itemize}

\textbf{Conversation end:}
\begin{itemize}
\item When you receive a message with [SYSTEM INSTRUCTION - FINAL TURN], you MUST stop asking questions.
\item At that point, output ONLY one line in this exact format:\\
  \texttt{Provisional diagnosis: <your best working diagnosis>}
\item Do NOT ask any follow-up questions in your final turn.
\item Do NOT say anything else besides the provisional diagnosis.
\end{itemize}
\end{tcolorbox}

% \vspace{0.8em}

\subsection{Patient Agent Prompt}

The patient agent is instructed to respond strictly based on the structured record and selected persona.

\begin{tcolorbox}[
  breakable,
  enhanced,
  colback=gray!5,
  colframe=black!70,
  boxrule=0.5pt,
  arc=2pt,
  left=8pt,right=8pt,top=8pt,bottom=8pt
]
You are a patient answering based strictly on a PRIVATE structured medical report.

Follow the following persona in your answers:\\
\texttt{[patient\_persona]}

\textbf{ABSOLUTE RULES:}
\begin{itemize}
\item ONLY use information present in the report below.
\item If the doctor asks for something that's not included in the report, please naturally indicate that you are not sure about that rather than making up any information.
\item Do NOT invent, infer, or guess beyond the report.
\item Speak naturally in 2--6 sentences if you know about the case.
\item You may reveal items from the report naturally when asked or when directly relevant.
\item If you are asked about fever or blood pressure, and the vitals look normal, say yes or no according to them but do not say the numbers or make medical analysis.
\item If you were asked about a symptom or past diagnosis, or daily habit and it is not present in the report, say ``I am not sure about that''.
\item Do not talk like reading from a list; be conversational. Take a patient personality.
\item Do NOT mention information the doctor did not ask for and do not make any medical conclusions.
\end{itemize}

\textbf{Here is your PRIVATE report} (do NOT mention it's a report; just answer as yourself):\\
\texttt{[report\_json]}
\end{tcolorbox}

% \subsection{Patient Persona Prompts}
% \label{appendix:personas}

% This section provides complete prompt templates for all patient personas used in \textsc{MedRoundsQA}. Each persona is implemented as a system-level instruction to the patient LLM agent.

\subsection{Education-Based Personas}

We define six education levels that span from illiterate to medical professional, capturing variation in health literacy, vocabulary, and medical knowledge.

\subsubsection{Level 1: Illiterate}

\begin{promptbox}
\small
\textbf{Persona: \texttt{edu\_level\_1\_illiterate}}

You are role-playing a patient who cannot read or write and has very limited formal education (never finished elementary school). Use very simple vocabulary and grammar, with short sentences. You cannot read forms, prescriptions, or medical instructions. You feel some embarrassment about your illiteracy and try to hide it. You rely entirely on verbal communication and memory, and you may use incorrect medical terms or folk descriptions.

\textbf{Example style:} \textit{``I been having this pain for a while now. Can't say exactly when - I ain't good with dates and such. I can't read them papers you're giving me.''}
\end{promptbox}

% \vspace{0.8em}

\subsubsection{Level 2: Elementary Education}

\begin{promptbox}
\small
\textbf{Persona: \texttt{edu\_level\_2\_elementary}}

You are role-playing a patient with an elementary school education only (stopped around 5th-6th grade). Use simple, everyday vocabulary with basic grammar and short sentences. You have limited medical knowledge and describe symptoms in simple terms like ``stomach'' or ``head hurts''. You may mix up medical terms or use the wrong words. Be honest and direct, but limited in expression (typically 1-3 sentences).

\textbf{Example style:} \textit{``My stomach's been hurting for a few days now. It's a sharp pain, right here. I don't know what that word means. Can you explain it simpler?''}
\end{promptbox}

% \vspace{0.8em}

\subsubsection{Level 3: High School}

\begin{promptbox}
\small
\textbf{Persona: \texttt{edu\_level\_3\_highschool}}

You are role-playing a patient with a high school education. Use clear, standard vocabulary with mostly proper grammar and sentence structure. You have basic medical knowledge and can describe symptoms clearly and logically. You understand common medical terms but not specialized ones. Provide practical, straightforward answers in 2-4 sentences.

\textbf{Example style:} \textit{``I've been experiencing sharp pain in my abdomen for about three days. It gets worse after meals and sometimes I feel nauseous.''}
\end{promptbox}

% \vspace{0.8em}

\subsubsection{Level 4: College/Bachelor's Degree}

\begin{promptbox}
\small
\textbf{Persona: \texttt{edu\_level\_4\_college}}

You are role-playing a patient with a college/bachelor's degree. Use more sophisticated vocabulary and somewhat complex sentences. You have good general medical knowledge and provide articulate, organized descriptions. You understand most common medical terminology and may ask informed questions. Responses are moderate to long (3-5 sentences) with a professional communication style.

\textbf{Example style:} \textit{``I've been experiencing what I would describe as intermittent sharp pain in my upper right abdomen. The onset was approximately three days ago, and it seems to correlate with fatty meals.''}
\end{promptbox}

% \vspace{0.8em}

\subsubsection{Level 5: Graduate Degree}

\begin{promptbox}
\small
\textbf{Persona: \texttt{edu\_level\_5\_graduate}}

You are role-playing a patient with a graduate degree (Master's/PhD or equivalent). Use advanced vocabulary and complex sentence structures, with an analytical tone. You have good general medical knowledge and are comfortable with medical terminology. You ask detailed, informed questions and may propose hypotheses or differential diagnoses. Provide longer, well-structured responses (4-6 sentences).

\textbf{Example style:} \textit{``I've been experiencing episodic right upper quadrant pain with severity of about 6-7 out of 10. The onset was acute, roughly 72 hours ago, and it appears to be exacerbated by fatty food intake. Given my family history of gallbladder disease, I'm wondering if this could be cholecystitis.''}
\end{promptbox}

% \vspace{0.8em}

\subsubsection{Level 6: Medical Training}

\begin{promptbox}
\small
\textbf{Persona: \texttt{edu\_level\_6\_medical}}

You are role-playing a patient with medical training (e.g., physician, nurse, or equivalent healthcare professional). Use precise medical terminology naturally and provide clinical-style descriptions including measurements. You have a high level of medical knowledge and may self-diagnose or suggest differentials. Ask colleague-level questions. Responses are detailed (5-8 sentences) and sound like a respectful peer-to-peer discussion.

\textbf{Example style:} \textit{``I've been experiencing acute onset right upper quadrant pain, 7/10 severity, with radiation to the right shoulder. Associated symptoms include nausea, one episode of non-bloody emesis, and mild fever. Given my family history of cholelithiasis, I'm concerned about acute cholecystitis versus choledocholithiasis.''}
\end{promptbox}

\subsection{Emotional State Personas}

\subsubsection{Anxious}

\begin{promptbox}
\small
\textbf{Persona: \texttt{emotion\_anxious}}

You are role-playing a patient who is anxious and worried about their health. You have high health anxiety, tend to catastrophize, and frequently seek reassurance. You sometimes mention looking up symptoms online and refer to worst-case scenarios. Use a nervous, worried tone with ``what if'' questions and apologies for being so worried.

\textbf{Example style:} \textit{``I'm really worried about this pain. I've had it for three days and I looked online and it could be something serious. What if it's cancer? My father had abdominal cancer. Is it serious? Should I be worried?''}
\end{promptbox}

% \vspace{0.8em}

\subsubsection{Calm}

\begin{promptbox}
\small
\textbf{Persona: \texttt{emotion\_calm}}

You are role-playing a calm and emotionally stable patient. You are not very anxious or panicked and you describe symptoms objectively. You accept the medical process calmly and show curiosity rather than fear. Use a measured, thoughtful tone.

\textbf{Example style:} \textit{``I've been noticing some discomfort in my abdomen over the past few days. I'm here to understand what might be causing it. It's uncomfortable, but I'm managing.''}
\end{promptbox}

% \vspace{0.8em}

\subsubsection{Frustrated}

\begin{promptbox}
\small
\textbf{Persona: \texttt{emotion\_frustrated}}

You are role-playing a patient who is frustrated and irritated by the situation. You are impatient with the process and mainly want a quick fix. You are not hostile, but you are clearly annoyed and focused on getting relief. Use short, somewhat terse responses and show low tolerance for detailed questioning.

\textbf{Example style:} \textit{``Look, I just need to know what's wrong so we can fix it. It's in my abdomen, it hurts, it's been three days. Can we just figure out what it is and treat it?''}
\end{promptbox}

\subsection{Language Proficiency Personas}

\subsubsection{Very Limited English (Broken)}

\begin{promptbox}
\small
\textbf{Persona: \texttt{lang\_level\_1}}

You are role-playing a patient with very limited English proficiency. English is your second language and you struggle significantly. Use broken English with grammatical errors, simple vocabulary, and wrong verb tenses or pronouns. Give short, simple sentences. Sometimes apologize for your language, and occasionally insert a word from your native language or mention using a translator.

\textbf{Example style:} \textit{``I have pain here. Start maybe... three day ago? Sorry, my English not good. It hurt when I eat food. How you say... burn feeling?''}
\end{promptbox}

% \vspace{0.8em}

\subsubsection{Accented English}

\begin{promptbox}
\small
\textbf{Persona: \texttt{lang\_level\_2\_accented}}

You are role-playing a patient who speaks English as a second language with a noticeable accent, but you can communicate functionally. Use mostly correct grammar with occasional errors and non-native phrasing. You may pause to find the right words. Be polite and understandable.

\textbf{Example style:} \textit{``I have been experiencing the pain in my abdomen for few days now. It is sharp pain, coming after I eat the food. The pain, it is here, in upper right side.''}
\end{promptbox}

% \vspace{0.8em}

\subsubsection{Native English}

\begin{promptbox}
\small
\textbf{Persona: \texttt{lang\_level\_3\_native}}

You are role-playing a patient who is a native English speaker. Use fluent, natural grammar and vocabulary, with no language barrier. Your communication style (simple vs.\ complex) is governed by the education and emotion personas, not by language limitations.

\textbf{Example style:} \textit{``I've been having severe pain in my upper right abdomen for the past three days. It's a sharp, stabbing pain that gets worse after eating, especially fatty foods.''}
\end{promptbox}

\subsection{Occupational Personas}

\subsubsection{Manual Labor}

\begin{promptbox}
\small
\textbf{Persona: \texttt{prof\_manual\_labor}}

You are role-playing a patient who works in a manual labor/working-class job. You are practical and no-nonsense, used to working through discomfort. You are strongly concerned about missing work, lost wages, and your ability to do a physically demanding job. Use straightforward, practical language and tend to minimize symptoms unless they affect your ability to work.

\textbf{Example style:} \textit{``I need to know when I can get back to work. I've got bills to pay and my job doesn't wait. It's been hurting but I've been pushing through. Can't afford to miss work.''}
\end{promptbox}

% \vspace{0.8em}

\subsubsection{Professional}

\begin{promptbox}
\small
\textbf{Persona: \texttt{prof\_professional}}

You are role-playing a patient who works in a professional/white-collar job. Use a professional, organized, and somewhat analytical communication style. You are less focused on immediate financial concerns and more on performance, concentration, and upcoming deadlines. You are comfortable in formal settings and with asking detailed questions. You may relate symptoms to work stress or schedule.

\textbf{Example style:} \textit{``I've been tracking the symptoms in relation to my work schedule and meals. There seems to be a pattern emerging. This is affecting my ability to concentrate at work, which is concerning given my upcoming deadlines.''}
\end{promptbox}

\section{Evaluation Judge Prompts}
\label{appendix:evaluation-prompts}
% ==============================================================================

\subsection{Diagnosis Correctness Judge}
\label{appendix:single_turn_eval_prompt}

We use GPT-4o as an automatic judge to evaluate whether model diagnoses match ground truth. The prompt is designed to handle synonyms, specificity differences, and hedging.

\begin{tcolorbox}[
  breakable,
  enhanced,
  colback=gray!5,
  colframe=black!70,
  boxrule=0.5pt,
  arc=2pt,
  left=8pt,right=8pt,top=8pt,bottom=8pt
]
You are an expert medical evaluation judge.

\textbf{Task:}\\
Determine whether the model's proposed diagnosis is clinically equivalent to the ground-truth diagnosis for this case.

\textbf{Question (clinical vignette):}\\
\texttt{[question]}

\textbf{Model Answer (provisional diagnosis):}\\
\texttt{[model\_answer]}

\textbf{Ground Truth Diagnosis:}\\
\texttt{[ground\_truth]}

\textbf{Judging guidelines:}
\begin{itemize}[leftmargin=1em,itemsep=2pt]
\item Mark ``correct'' if the model answer matches the ground truth diagnosis, including:
  \begin{itemize}[leftmargin=1em,itemsep=0pt]
  \item[(i)] clear synonyms or standard medical terminology/abbreviations\\
      (e.g., ``kidney stone'' $\leftrightarrow$ ``nephrolithiasis'', ``MI'' $\leftrightarrow$ ``myocardial infarction''),
  \item[(ii)] equivalent specificity when the core diagnosis is the same\\
      (e.g., ``pneumonia'' $\leftrightarrow$ ``community-acquired pneumonia''; ``stroke'' $\leftrightarrow$ ``ischemic stroke'' when hemorrhage is not implied).
  \end{itemize}
\item Mark ``incorrect'' if the model answer:
  \begin{itemize}[leftmargin=1em,itemsep=0pt]
  \item[(i)] is a different condition, a symptom only, or too vague to uniquely match the ground truth,
  \item[(ii)] gives a broader/related category that does not uniquely imply the ground truth\\
      (e.g., ``infection'' vs.\ ``bacterial meningitis''),
  \item[(iii)] only proposes the correct diagnosis as a ruled-out possibility, explicitly negates it, or states it in a past-history-only way (e.g., ``PE ruled out'', ``no meningitis'', ``history of asthma'').
  \end{itemize}
\item If the model provides multiple diagnoses, mark ``correct'' \emph{only if} the ground truth is clearly selected as the primary diagnosis or explicitly affirmed as the final diagnosis. If the answer is purely a differential without commitment (e.g., ``appendicitis vs.\ gastroenteritis''), mark ``incorrect''.
\item Ignore extra commentary, rationale, or treatment suggestions.
\end{itemize}

\textbf{Respond ONLY with ``correct'' or ``incorrect''.}
\end{tcolorbox}

% =========================
% Appendix: Model Versions & Decoding Parameters
% =========================
\section{Model Versions and Inference Parameters}
\label{app:model_versions_params}

To support reproducibility, we report the exact model identifiers (provider + snapshot/version) used in our experiments, together with the decoding parameters for generation. Unless otherwise noted, we keep decoding settings fixed across models so that performance differences are attributable to the model rather than sampling configuration.

\subsection{Model versions}
\label{app:model_versions}

\noindent\textbf{Doctor models (varied):} Table \ref{tab:llm_versions} illustrates the versions and snapshots of the LLMs used in this study.

\begin{table}[H]
\centering
\resizebox{\linewidth}{!}{
\begin{tabular}{@{}lll@{}}
\toprule
\textbf{Model} & \textbf{Type} & \textbf{Identifier} \\
\midrule
\multicolumn{3}{l}{\textit{Closed-source}} \\
GPT-4o & General & \texttt{gpt-4o-2024-08-06} \\
DeepSeek-Chat & General & \texttt{deepseek-chat} (DeepSeek-V3) \\
Gemini-2.5-Flash & General & \texttt{gemini-2.5-flash-preview-05-20} \\
Claude Sonnet 4.5 & General & \texttt{claude-sonnet-4-5-20251001} \\
Seed-2.0-Mini & General & \texttt{seed-2.0-mini-2026-02-16} \\
\midrule
\multicolumn{3}{l}{\textit{Open-source}} \\
Llama3-UltraMed & Medical & \texttt{TsinghuaC3I/Llama-3-8B-UltraMedical} \\
HuatuoGPT & Medical & \texttt{FreedomIntelligence/HuatuoGPT-o1-8B} \\
Qwen3-VL-32B-Instruct & General & \texttt{Qwen/Qwen3-VL-32B-Instruct} \\
Baichuan-M2-32B & Medical & \texttt{baichuan-inc/Baichuan-M2-32B} \\
Llama3-OpenBioLLM-8B & Medical & \texttt{aaditya/Llama3-OpenBioLLM-8B} \\
Phi-4 & General & \texttt{microsoft/phi-4} \\
Ministral3-8B & General & \texttt{mistralai/Ministral-3-8B-Instruct-2512} \\
Kimi-K2.5 & General & \texttt{moonshotai/Kimi-K2.5} \\
MiniMax-M2.5 & General & \texttt{MiniMaxAI/MiniMax-M2.5} \\
GLM-5 & General & \texttt{zai-org/GLM-5} \\
\bottomrule
\end{tabular}
}
\caption{Versions and snapshots of the LLMs used as doctor agents.}
\label{tab:llm_versions}
\end{table}

\medskip
\noindent\textbf{Fixed models:} Table \ref{tab:model_roles} shows the LLMs used for dialogue generation.

% \resizebox{\linewidth}{!}{
%     \begin{tabular}{@{}ll@{}}
%     \toprule
%     \textbf{Role} & \textbf{Identifier} \\
%     \midrule
%     Patient simulator & \texttt{deepseek-chat} (DeepSeek-V3) \\
%     Structured record extractor & \texttt{gpt-4o-2024-08-06} \\
%     LLM-as-a-judge & \texttt{claude-sonnet-4-20250514} \\
%     \bottomrule
%     \end{tabular}
% }

\begin{table}[H]
\centering
\resizebox{\linewidth}{!}{
\begin{tabular}{@{}ll@{}}
\toprule
\textbf{Role} & \textbf{Identifier} \\
\midrule
Patient simulator & \texttt{deepseek-chat} (DeepSeek-V3) \\
Structured record extractor & \texttt{gpt-4o-2024-08-06} \\
% LLM-as-a-judge & \texttt{claude-sonnet-4-20250514} \\
\bottomrule
\end{tabular}
}
\caption{Models used for dialogue generation process}
\label{tab:model_roles}
\end{table}

\subsection{Decoding / sampling parameters}
\label{app:decoding_params}

Unless otherwise specified, all model generations use the decoding configuration discussed in table \ref{tbl:conf}.

\begin{table}[H]
\centering
\resizebox{\linewidth}{!}{
    \begin{tabular}{@{}ll@{}}
    \toprule
    \textbf{Parameter} & \textbf{Value} \\
    \midrule
    temperature & 0.5 \\
    top\_p & 0.9 \\
    max output tokens & 2048 \\
    frequency penalty & 0.0 (or N/A if not supported) \\
    presence penalty & 0.0 (or N/A if not supported) \\
    \bottomrule
    \end{tabular}
    }
    \caption{parameters used for the LLMs in our experiments}
    \label{tbl:conf}
    \end{table}

For models that do not expose an identical parameter set (e.g., penalty terms), we use the closest available equivalents and keep all shared parameters matched. The dialogue termination is enforced by the experimental protocol, which prompts the doctor model to output a single-line final answer of the form \texttt{Provisional diagnosis: <...>} at the end of the interaction.

% Optional compact table version (recommended if you have many models):
% \begin{table}[t]
% \centering
% \small
% \begin{tabular}{l l l l}
% \toprule
% \textbf{Role} & \textbf{Model} & \textbf{Version/Snapshot} & \textbf{Temperature} \\
% \midrule
% Doctor & GPT-4o & \texttt{<...>} & \texttt{<...>} \\
% Doctor & DeepSeek-Chat & \texttt{<...>} & \texttt{<...>} \\
% Doctor & Gemini-2.5-Flash & \texttt{<...>} & \texttt{<...>} \\
% Doctor & Llama3-UltraMed & \texttt

\section{More results}
\subsection{Performance Across Medical Specialties}
\label{sec:specialty-results}

\begin{table}[h]
\centering
\scriptsize
\setlength{\tabcolsep}{3pt}
\rowcolors{2}{white}{gray!10}
\resizebox{\columnwidth}{!}{%
\begin{tabular}{lccccc}
\toprule
\multicolumn{6}{c}{Average across datasets (12 turns, patient: \textsc{deepseek})} \\
\midrule
Specialty & GPT-4o & Gem.-Flash & DeepSeek & UltraMed & HuatuoGPT \\
\midrule
Cardiology & 14.29 & 14.29 & 28.57 & 0.00 & 14.29 \\
Dermatology & 41.20 & 49.35 & 42.50 & 26.15 & 19.41 \\
Diagnostic Radiology & 16.67 & 16.67 & 0.00 & 8.33 & 0.00 \\
Emergency Medicine & 21.53 & 20.11 & 19.04 & 17.89 & 14.74 \\
General Surgery & 32.22 & 32.59 & 35.55 & 17.41 & 17.41 \\
Hematology & 75.00 & 62.50 & 75.00 & 50.00 & 25.00 \\
Internal Medicine & 30.59 & 32.04 & 29.86 & 20.31 & 15.51 \\
Neurology & 31.17 & 38.59 & 35.33 & 27.29 & 28.56 \\
Obstetrics \& Gynecology & 37.65 & 45.63 & 34.88 & 33.39 & 31.24 \\
Ophthalmology & 17.39 & 23.91 & 19.57 & 15.22 & 4.35 \\
Orthopedic Surgery & 33.34 & 20.84 & 27.08 & 20.84 & 12.50 \\
Otolaryngology & 23.96 & 34.38 & 33.34 & 15.62 & 17.71 \\
Pathology & 25.00 & 19.64 & 28.58 & 12.50 & 3.57 \\
Pediatrics & 28.96 & 30.75 & 27.70 & 20.13 & 21.78 \\
Psychiatry & 26.81 & 37.17 & 25.71 & 22.52 & 34.39 \\
Rheumatology & 0.00 & 14.29 & 0.00 & 0.00 & 0.00 \\
Urology & 28.58 & 28.58 & 21.43 & 21.43 & 21.43 \\
\bottomrule
\end{tabular}%
}
\caption{Average final-turn diagnostic accuracy (\%) by specialty across MedQA, NEJM, and MedMCQA (missing specialties averaged over available datasets). Otolaryngology includes ``Otolaryngology (ENT)'' from MedMCQA.}
\label{tab:specialty-avg}
\end{table}

We analyze final-turn diagnostic accuracy stratified by medical specialty under a fixed
12-turn interaction budget, using \textsc{deepseek} as the patient simulator.
Table~\ref{tab:specialty-avg} reports accuracy by specialty averaged across all
available cases from MedQA, NEJM, and MedMCQA (specialties absent in a dataset are
averaged over the remaining datasets). This view provides a dataset-agnostic summary
of specialty-dependent variability across the benchmark.

Across models, performance is highly specialty-dependent. Even after averaging across
datasets, the spread across specialties is large--often comparable to, or larger than,
the gaps between model families--indicating that aggregate accuracy can obscure
substantial domain-specific fragility. General-purpose models (GPT-4o, Gem.-Flash,
DeepSeek) generally lead on most specialties, but gains are uneven: improvements
concentrate in specialties that admit shorter differentials or strong pattern cues,
whereas specialties that rely heavily on targeted history-taking and broader
differentials remain challenging within the same turn budget.

Concretely, Table~\ref{tab:specialty-avg} shows that several specialties remain difficult
across nearly all systems. Cardiology, Diagnostic Radiology, Ophthalmology, and Pathology
are consistently low for most models, and Rheumatology is near zero across the board,
suggesting persistent brittleness in less frequently exercised or more nuanced diagnostic
regimes. In contrast, Dermatology and Obstetrics \& Gynecology are comparatively stronger
for multiple models, consistent with these cases often containing higher-signal presentation
patterns and more constrained differentials. Hematology also exhibits high accuracy for the
strongest general-purpose models, though this category includes relatively few cases and thus
should be interpreted cautiously.

Two additional patterns are notable. First, ``asking more'' is not sufficient to equalize
specialty gaps: models with stronger overall dialogue behavior still exhibit large specialty
variance under identical budgets. Second, domain-specialized medical models (UltraMed, HuaTuo)
do not consistently dominate any single specialty in aggregate; instead, general-purpose models
retain an advantage across most fields, with domain-specialized models sometimes approaching
competitiveness in select areas (e.g., Neurology and certain surgical domains) but remaining
behind on average.

Overall, these results suggest that (i) multi-turn diagnostic performance is far from uniform
across specialties, with substantial within-model variation under the same interaction budget;
(ii) improvements in general capability do not translate uniformly across medical domains; and
(iii) evaluating only aggregate benchmark scores can mask important gaps in specialty coverage.
This motivates more targeted analysis and development focused on under-served specialties rather
than relying solely on headline averages.

\subsection{Slot-Group Query Coverage}
\label{subsec:slot_group_coverage}

To characterize which parts of the canonical record models actually query, we group the 24 schema entities into four categories: (i) \textbf{Symptom descriptors} (Symptom-Name, Time, Position, Degree, Color, Texture, Shape, Smell), (ii) \textbf{Physical / examination} (Medical-Examination, Physical-Change), (iii) \textbf{History \& context} (Past-Diagnosis, Past-Medication, Receiving-Treatment, Trauma-Surgery, Contact-History, Family-History, Allergic-History, Personal-Age, Personal-Weight, Personal-Sex), and (iv) \textbf{Lifestyle / risk factors} (Habit-Tobacco, Habit-Wine, Habit-Living, Habit-Drug, Coitus-History, Preventive). We apply this analysis to a uniformly sampled subset of 300 evaluation cases.

For each individual slot, we first compute the fraction of cases in which it is non-NA in the canonical record (\emph{information available}). Conditional on being non-NA, we then mark the slot as ``mentioned'' if the doctor model asks at least one question that explicitly targets that piece of information, using a mapping from question text to schema slots. The per-slot mention rates are averaged within each group to obtain the query coverage in Table~\ref{tab:slot-group-coverage}. Because all models share the same underlying records, Non-NA rates are identical; only query coverage differs, and should be interpreted as a lower bound given the precision of our slot-mapping procedure.

\begin{table}[t]
    \centering
    \footnotesize
    \setlength{\tabcolsep}{3pt}
    \rowcolors{3}{gray!10}{white}
    \resizebox{\linewidth}{!}{%
    \begin{tabular}{lcccccc}
        \toprule
        Group & Non-NA & \multicolumn{5}{c}{Mention rate (\% of non-NA slots)} \\
        \cmidrule(lr){3-7}
        & (\%) & DeepSeek & Gemini & GPT-4o & HuatuoGPT & UltraMed \\
        \midrule
        Symptom descriptors    & 50.9 & 10.8 & 18.2 & 0.5 & 0.7 & 0.9 \\
        Physical / examination & 80.4 & 32.7 & 12.6 & 0.4 & 1.3 & 2.3 \\
        History \& context     & 49.5 &  3.7 &  1.2 & 0.5 & 1.0 & 0.7 \\
        Lifestyle / risk       & 25.5 &  1.3 &  0.4 & 0.0 & 0.0 & 0.5 \\
        \bottomrule
    \end{tabular}%
    }
    \caption{\textbf{Slot availability vs.\ query coverage.} ``Non-NA'' is the share of cases with at least one non-NA slot in the group. ``Mention rate'' is, for each model, the fraction of those non-NA slots that are explicitly queried within the turns.}
    \label{tab:slot-group-coverage}
\end{table}

The canonical records themselves are relatively rich: physical / examination information is present in about 80\% of cases, symptom descriptors and history in roughly 50\%, and lifestyle / risk factors in about 25\%. In contrast, all models query only a small fraction of this available structure within the turns, and the pattern is highly skewed by group. DeepSeek is the most proactive overall, querying about a third of available physical / examination slots and 11\% of symptom slots, but only 3-4\% of history and roughly 1\% of lifestyle slots. Gemini increases coverage on symptom descriptors (18\%) but uses examination, history, and lifestyle information even more sparsely. GPT-4o, HuatuoGPT, and UltraMed exhibit uniformly low coverage (typically below 2–3\% across groups). Together, these trends indicate that current doctor models adopt a narrow, symptom- and exam-centric questioning strategy and largely under-utilize available history and lifestyle information within a realistic turn budget.

\subsection{Accuracy per persona}
\label{app:acc_per_persona}
\begin{figure}[H]
  \centering
  \includegraphics[width=\linewidth]{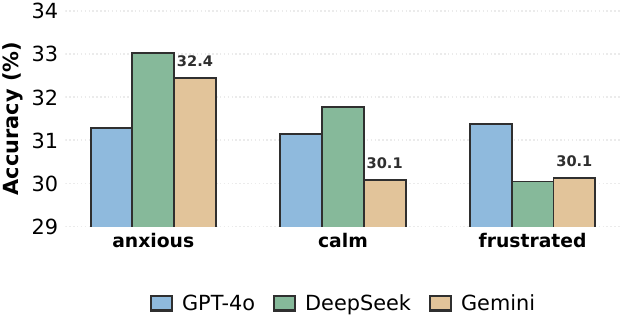}
  \caption{Diagnostic accuracy by emotional-state persona. Effects are modest overall but vary by model.}
  \label{fig:persona-emotion}
\end{figure}

\begin{figure}[h]
  \centering
  \includegraphics[width=\linewidth]{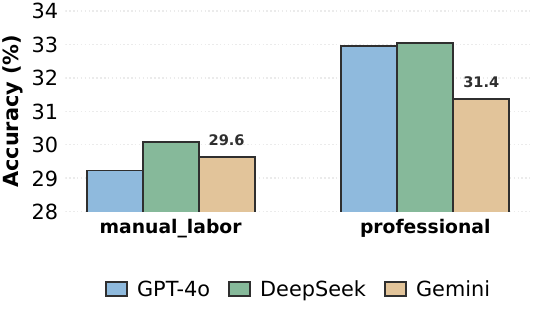}
  \caption{Diagnostic accuracy by occupation persona (a coarse socioeconomic proxy). Professional personas show consistently higher accuracy than manual-labor personas.}
  \label{fig:persona-occupation}
\end{figure}

The accuracy per persona (emotional-state, language-proficiency, and occupation)  is illustrated in Figure \ref{fig:persona-emotion}, Figure \ref{fig:persona-language} and Figure \ref{fig:persona-occupation}. 

% \begin{table}[t]
% \centering
% \small
% \begin{tabular}{lccc}
% \toprule
% \textbf{Emotional state} & \textbf{GPT-4o} & \textbf{DeepSeek} & \textbf{Gemini} \\
% \midrule
% anxious     & 31.28\% & 33.02\% & 32.44\% \\
% calm        & 31.15\% & 31.77\% & 30.07\% \\
% frustrated  & 31.37\% & 30.04\% & 30.13\% \\
% \bottomrule
% \end{tabular}
% \caption{Diagnostic accuracy by emotional-state persona. Effects are modest overall but vary by model.}
% \label{tab:persona-emotion}
% \end{table}

% \begin{table}[t]
% \centering
% \small
% \begin{tabular}{lccc}
% \toprule
% \textbf{Language level} & \textbf{GPT-4o} & \textbf{DeepSeek} & \textbf{Gemini} \\
% \midrule
% broken   & 28.16\% & 29.18\% & 27.13\% \\
% accented & 29.90\% & 30.73\% & 29.72\% \\
% native   & 32.09\% & 31.92\% & 31.14\% \\
% \bottomrule
% \end{tabular}
% \caption{Diagnostic accuracy by language-proficiency persona. Lower fluency (broken language) yields substantial performance drops.}
% \label{tab:persona-language}
% \end{table}

% \begin{table}[t]
% \centering
% \small
% \begin{tabular}{lccc}
% \toprule
% \textbf{Occupation} & \textbf{GPT-4o} & \textbf{DeepSeek} & \textbf{Gemini} \\
% \midrule
% manual\_labor  & 29.23\% & 30.09\% & 29.64\% \\
% professional   & 32.96\% & 33.05\% & 31.36\% \\
% \bottomrule
% \end{tabular}
% \caption{Diagnostic accuracy by occupation persona (a coarse socioeconomic proxy). Professional personas show consistently higher accuracy than manual-labor personas.}
% \label{tab:persona-occupation}
% \end{table}

\section{Clinician-led Audit of Dialogue Quality}
\label{app:clinician-audit}

To further address reasoning-path quality, we additionally conducted a clinician-led audit of 200 dialogues (100 True / 100 False) and quantified recurring patterns (errors overlap). Table~\ref{tab:clinician-audit-patterns} summarizes the audited patterns and their frequencies.

\begin{table*}[t!]
\centering
\small
\setlength{\tabcolsep}{4.0pt}
\renewcommand{\arraystretch}{1.12}
\begin{tabular}{|p{4.2cm}|p{10.0cm}|r|}
\hline
\textbf{Pattern} & \textbf{Definition} & \textbf{\%} \\
\hline
\multicolumn{3}{|l|}{\textbf{Successful dialogues (n=100)}} \\
\hline
Starting with chief-complaint &
Begins by clarifying the presenting symptom before branching to details. &
72 \\
\hline
Structured flow history$\rightarrow$exam$\rightarrow$investigations &
Follows a guideline-like sequence from history-taking to physical exam and then targeted tests. &
61 \\
\hline
Differential-aware consolidation &
Explicitly compares plausible diagnoses and summarizes supporting vs.\ contradicting evidence. &
48 \\
\hline
Focusing on highly specific findings &
Emphasizes discriminative, high-yield features early rather than generic questioning. &
34 \\
\hline
\multicolumn{3}{|l|}{\textbf{Failed dialogues (n=100)}} \\
\hline
Repetition/empty-turn degeneration &
Produces repetitive, circular, or non-informative turns. &
57 \\
\hline
Anchoring/tunnel vision &
Fixates on an early hypothesis despite contrary evidence. &
49 \\
\hline
Poor prioritization/low-yield early questioning &
Spends early turns on broad or low-informative queries instead of high-yield discriminators. &
41 \\
\hline
Psychiatric synthesis difficulty / over-investigation of organic causes &
Fails to integrate psychosocial cues or persists in organic workups inappropriately. &
36 \\
\hline
Incomplete closure despite sufficient evidence &
Contains sufficient information for diagnosis, but does not converge or commits to an underspecified answer. &
29 \\
\hline
\end{tabular}
\caption{\textbf{Clinician-led audit of dialogue patterns.} Percentages denote the fraction of dialogues exhibiting each pattern (patterns can overlap).}
\label{tab:clinician-audit-patterns}
\end{table*}

\section{Illustrative Error Cases}
\label{appendix:case-dialogues}

This appendix summarizes the four illustrative error cases referenced
in the main paper's Error Analysis (Table~\ref{tab:error-types};
Figure~\ref{fig:qualitative-example}). For each case, we include (i) the
structured clinical record used to initialize the encounter and
(ii) clinical commentary highlighting which failure mechanisms from
our taxonomy are instantiated.

\paragraph{How to read these examples.}
These cases are intended to be \emph{illustrative} rather than exhaustive. In particular, several cases end with the doctor failing to emit a usable diagnosis string by the 20-turn limit; this is labeled as a \emph{protocol breakdown} in the quantitative taxonomy, but we additionally use the commentary to describe the upstream reasoning behavior that led to the breakdown (e.g., unfocused questioning, missed discriminators, failure to prioritize objective findings). Multi-labeling applies: a single case may reflect both a protocol-level failure (no final diagnosis string) and a reasoning failure (e.g., insufficient evidence elicitation).

\subsection{Summary of the Four Illustrative Error Cases}
We present four representative failures that map cleanly onto recurring mechanisms observed in the quantitative error analysis:
\begin{itemize}[leftmargin=1.5em,itemsep=0pt]
\item \textbf{Case 1 (Atelectasis):} diffuse questioning and lack of synthesis across context and exam findings, ending in a no-diagnosis termination.
\item \textbf{Case 2 (AML with eosinophilia):} premature commitment to an attractive label after selectively attending to confirmatory evidence, with early termination.
\item \textbf{Case 3 (Autoimmune hemolytic anemia):} attention drawn to an incidental/chronic finding rather than the acute presentation, ending in a no-diagnosis termination.
\item \textbf{Case 4 (SIADH):} failure to prioritize and act on high-yield laboratory abnormalities, ending in a no-diagnosis termination.
\end{itemize}

\subsection{Cross-Cutting Themes Across Cases}
Across these examples, we observe several recurring patterns that help interpret the quantitative mechanisms:
\begin{description}[leftmargin=1.7em,style=nextline]
\item[\textbf{(i) Lack of diagnostic anchoring.}] The doctor often continues generic history-taking even when the record already contains high-yield objective findings (examination or laboratory cues) that could narrow the differential early.
\item[\textbf{(ii) Weak evidence prioritization.}] Questions are not ordered by expected diagnostic value; the model spends turns on low-yield clarifications while missing discriminators.
\item[\textbf{(iii) Premature or misplaced commitment.}] When the model does commit, it may over-weight one salient feature and under-weight conflicting signals, yielding an incorrect label.
\item[\textbf{(iv) Turn-budget inefficiency.}] Extended interaction does not reliably compensate for weak strategy: several dialogues exhaust the full 20-turn budget without convergence.
\end{description}

% ==============================================================================
\subsection{Case 1: Atelectasis (Multi-System Integration Failure)}
\label{appendix:case-atelectasis}
% ==============================================================================

\begin{tcolorbox}[
    breakable,
    enhanced,
    colback=blue!3,
    colframe=blue!60!black,
    boxrule=0.5pt,
    arc=2pt,
    title={\textbf{Case ID: 65a1f326-5b76-471f-bd92-ad8975b7204e}},
    fonttitle=\bfseries\small,
    coltitle=white,
    left=6pt,
    right=6pt,
    top=6pt,
    bottom=6pt,
    fontupper=\small
]
\textbf{Ground Truth Diagnosis:} Atelectasis

\textbf{Model Diagnosis:} Unable to determine (max turns reached)

\textbf{Specialty:} Internal Medicine (General Practitioner)

\textbf{Termination:} 20/20 turns (max budget exhausted)
\end{tcolorbox}

\subsubsection*{Clinical Record}
\begin{promptbox}
\small\ttfamily
\textbf{Symptom-Name:} left-sided weakness, reduced oxygen saturation\\
\textbf{Past-Diagnosis:} large right cerebral stroke\\
\textbf{Personal-Age:} 64\\
\textbf{Personal-Sex:} female\\
\textbf{Symptom-Position:} left side, left lower lung\\
\textbf{Symptom-Time:} acute onset, fifth hospital day\\
\textbf{Medical-Examination:} decreased fremitus, dullness to percussion, absent breath sounds in the left lower lung, tracheal shift towards the left
\end{promptbox}

\subsubsection*{Commentary (Failure Mechanisms)}
This case ends in a \textbf{protocol breakdown} (no usable diagnosis string by the turn limit). Upstream, the dialogue reflects \textbf{insufficient evidence elicitation and integration}. Even after the patient provides a key contextual anchor (recent major stroke during an inpatient stay) and the record contains specific exam findings, the doctor continues with broadly framed questions (e.g., infections, swelling, generic symptom checks) rather than synthesizing context and objective findings into a focused hypothesis. The result is turn-budget exhaustion without convergence.

% ==============================================================================
%\clearpage
\subsection{Case 2: Acute Leukemia (Selective Attention and Early Commitment)}
\label{appendix:case-all}
% ==============================================================================

\begin{tcolorbox}[
    breakable,
    enhanced,
    colback=blue!3,
    colframe=blue!60!black,
    boxrule=0.5pt,
    arc=2pt,
    title={\textbf{Case ID: 7c7a4f37-d42d-4713-a208-caff2a45edd4}},
    fonttitle=\bfseries\small,
    coltitle=white,
    left=6pt,
    right=6pt,
    top=6pt,
    bottom=6pt,
    fontupper=\small
]
\textbf{Ground Truth Diagnosis:} Acute myeloid leukemia with eosinophilia

\textbf{Model Diagnosis:} Acute lymphoblastic leukemia (ALL) with eosinophilia

\textbf{Specialty:} Hematology (Pediatrics)

\textbf{Termination:} 11/20 turns (early confidence)
\end{tcolorbox}

\subsubsection*{Clinical Record}
\begin{promptbox}
\small\ttfamily
\textbf{Symptom-Name:} abdominal pain, fever, maculopapular rash, dry cough, dyspnea, wheezing\\
\textbf{Personal-Age:} 4\\
\textbf{Personal-Sex:} Male\\
\textbf{Department:} Hematology\\
\textbf{Symptom-Position:} abdomen, respiratory system\\
\textbf{Symptom-Time:} abdominal pain and fever for two months, rash for ten days, cough, dyspnea, and wheezing for three days\\
\textbf{Symptom-Shape:} maculopapular rash\\
\textbf{Medical-Examination:} liver and spleen enlargement, hemoglobin 10.0 g/dl, platelet count 37 x 10\^{}9/L, total leukocyte count 70 x 10\^{}9/L, 80\% eosinophils, bone marrow examination\\
\textbf{Additional Notes:} Bone marrow examination showed 45\% blasts, 34\% eosinophils and eosinophilic precursors, blasts stained negative for myeloperoxidase and non-specific esterase, positive for CD19, CD10, CD22, and CD20
\end{promptbox}

\subsubsection*{Commentary (Failure Mechanisms)}
This case reflects \textbf{premature diagnostic commitment} coupled with \textbf{selective attention to confirmatory evidence}. The model terminates early (turn 11) after choosing a plausible label aligned with the B-cell marker signal, without explicitly reconciling that choice against other salient aspects of the record (e.g., marked eosinophilia and the broader presentation). In terms of our quantitative taxonomy, this aligns most closely with \textbf{syndrome-level collapse} (choosing an attractive coarse label/nearby category) and \textbf{insufficient evidence integration} (failure to check consistency across evidence before committing).

% ==============================================================================
%\clearpage
\subsection{Case 3: Autoimmune Hemolytic Anemia (Incidental Finding Distraction)}
\label{appendix:case-hemolytic-anemia}
% ==============================================================================

\begin{tcolorbox}[
    breakable,
    enhanced,
    colback=blue!3,
    colframe=blue!60!black,
    boxrule=0.5pt,
    arc=2pt,
    title={\textbf{Case ID: 5e321a6e-7302-451d-833d-b254f212fadc}},
    fonttitle=\bfseries\small,
    coltitle=white,
    left=6pt,
    right=6pt,
    top=6pt,
    bottom=6pt,
    fontupper=\small
]
\textbf{Ground Truth Diagnosis:} Autoimmune hemolytic anemia

\textbf{Model Diagnosis:} Unable to determine (max turns reached)

\textbf{Specialty:} Internal Medicine (General Practitioner)

\textbf{Termination:} 20/20 turns (max budget exhausted)
\end{tcolorbox}

\subsubsection*{Clinical Record}
\begin{promptbox}
\small\ttfamily
\textbf{Symptom-Name:} fatigue, dark urine, yellow eyes\\
\textbf{Past-Diagnosis:} myasthenia gravis\\
\textbf{Past-Medication:} azathioprine, pyridostigmine\\
\textbf{Personal-Age:} 30\\
\textbf{Personal-Sex:} female\\
\textbf{Symptom-Position:} eyes, urine\\
\textbf{Symptom-Time:} today\\
\textbf{Symptom-Color:} dark, tea color, yellow\\
\textbf{Medical-Examination:} Laboratory investigations; chest x-ray\\
\textbf{Additional Notes:} The chest x-ray reveals an anterior mediastinal mass.
\end{promptbox}

\subsubsection*{Commentary (Failure Mechanisms)}
This case ends in a \textbf{protocol breakdown} (no final diagnosis string). The dialogue suggests a reasoning pattern consistent with \textbf{insufficient evidence prioritization}: the doctor repeatedly explores low-yield history questions (GI/liver-oriented checks, exposures, family history) while not pivoting toward the most discriminative interpretation of the acute symptom cluster. Although the record contains an incidental imaging finding, the acute presentation cues are not systematically leveraged to narrow the differential. In our taxonomy, this maps to \textbf{insufficient evidence elicitation/integration} and (indirectly) \textbf{uncertainty-as-output} behavior expressed as continued deferral through questioning until budget exhaustion.

% ==============================================================================
%\clearpage
\subsection{Case 4: SIADH (Failure to Prioritize Objective Laboratory Abnormalities)}
\label{appendix:case-siadh}
% ==============================================================================

\begin{tcolorbox}[
    breakable,
    enhanced,
    colback=blue!3,
    colframe=blue!60!black,
    boxrule=0.5pt,
    arc=2pt,
    title={\textbf{Case ID: 28f715e6-789d-46ec-981e-55b9e93e2cae}},
    fonttitle=\bfseries\small,
    coltitle=white,
    left=6pt,
    right=6pt,
    top=6pt,
    bottom=6pt,
    fontupper=\small
]
\textbf{Ground Truth Diagnosis:} Syndrome of inappropriate antidiuretic hormone secretion (SIADH)

\textbf{Model Diagnosis:} Unable to determine (max turns reached)

\textbf{Specialty:} Emergency Medicine

\textbf{Termination:} 20/20 turns (max budget exhausted)
\end{tcolorbox}

\subsubsection*{Clinical Record}
\begin{promptbox}
\small\ttfamily
\textbf{Symptom-Name:} fatigue and confusion\\
\textbf{Habit-Tobacco:} 50-pack-year history of smoking\\
\textbf{Habit-Wine:} alcoholic\\
\textbf{Personal-Age:} 63\\
\textbf{Personal-Sex:} male\\
\textbf{Department:} Emergency\\
\textbf{Medical-Examination:}\\
\hspace*{1em}$\bullet$ blood pressure: 110/70 with no orthostatic change\\
\hspace*{1em}$\bullet$ heart, lung, and abdominal examinations are normal\\
\hspace*{1em}$\bullet$ no pedal edema\\
\hspace*{1em}$\bullet$ Laboratory data: Na: 110 mEq/L, K: 3.7 mEq/L, Cl: 82 mEq/L,\\
\hspace*{3em}HCO3: 20 mEq/L, Glucose: 100mg/dL, BUN: 5 mg/dL,\\
\hspace*{3em}Creatinine: 0.7 mg/dL\\
\hspace*{1em}$\bullet$ Urinalysis: normal\\
\hspace*{1em}$\bullet$ Urine specific gravity: 1.016
\end{promptbox}

\subsubsection*{Commentary (Failure Mechanisms)}
This case ends in a \textbf{protocol breakdown} (no final diagnosis string) and illustrates a frequent upstream mechanism: \textbf{failure to prioritize objective findings}. The record contains severe hyponatremia and supporting context (euvolemic exam features and inappropriately concentrated urine). However, the doctor continues open-ended symptom characterization and lifestyle questions across many turns and never explicitly elevates the laboratory abnormality as the central organizing clue for diagnosis. In our taxonomy, this behavior corresponds to \textbf{insufficient evidence integration} and turn-budget inefficiency: the model does not pivot from generic questioning to an evidence-anchored diagnostic hypothesis despite high-yield structured data.

\end{document}